%% file: main.tex
\documentclass[11pt]{article}

\usepackage[preprint]{acl}

\usepackage{times}
\usepackage{latexsym}

\usepackage[T1]{fontenc}

\usepackage[utf8]{inputenc}

\usepackage{microtype}

\usepackage{inconsolata}

\usepackage{graphicx}
\usepackage{subcaption}
\usepackage{soul}
\usepackage{enumitem}

\usepackage[most]{tcolorbox}

\newtcolorbox{promptbox}[1][]{
  breakable,
  colback=gray!5,
  colframe=gray!50!black,
  fonttitle=\bfseries,
  title=Prompt,
  boxrule=0.5pt,
  arc=2pt,
  left=6pt, right=6pt, top=4pt, bottom=4pt,
  #1
}

\usepackage{tabularx}
\usepackage{xltabular}
\usepackage{multirow}
\usepackage{float}
\usepackage{booktabs}

\title{Curiosity as Linguistic Intervention: Using LLM Tutoring Dialogues to Influence Exploratory Learning Behavior}

\author{Gevindu Ganganath\protect\textsuperscript{1}, Pasindu Bolonghege\protect\textsuperscript{1}, Qianru Lyu\protect\textsuperscript{2}, Pradeep Varakantham\protect\textsuperscript{1},\\
\textbf{Thivya Kandappu\protect\textsuperscript{1}} \\
\protect\textsuperscript{1}School of Computing and Information Systems, Singapore Management University \\
\protect\textsuperscript{2}Human-Computer Interaction Institute, Carnegie Mellon University \\
\texttt{gevindu.wk.2025@phdcs.smu.edu.sg, pasindub@smu.edu.sg, qlyu@andrew.cmu.edu}\\
\texttt{pradeepv@smu.edu.sg, thivyak@smu.edu.sg}}
  
\newcommand{\system}{\textsc{CurioBot }}
\newcommand{\systemnospace}{\textsc{CurioBot}}

\newbool{showComments}
\setbool{showComments}{false}  
\ifbool{showComments}{

}

\begin{document}
\maketitle

\begin{abstract}
Large Language Models (LLMs) provide a new opportunity to study how language shapes exploratory cognition because conversational strategies can be systematically manipulated at inference time. We introduce \systemnospace, a framework that operationalizes Berlyne's collative variables, novelty, complexity, conflict, and uncertainty, as adaptive linguistic interventions for conversational tutoring. Across 270 tutoring conversations spanning multiple model families, domains, and topic complexity levels, curiosity-oriented interventions consistently increased exploratory learner behaviors, producing up to 2.4$\times$ more conversational turns under fixed time budgets. To measure these effects, we further introduce a learner-centered evaluation framework capturing exploratory questioning, conversational agency, productive struggle, and observable curiosity. Learner-side gains persisted even when tutor-side instructional quality remained unchanged, suggesting that curiosity functions as a partially independent interaction-level mechanism. More broadly, our results demonstrate that LLM-mediated dialogue can serve as a scalable experimental framework for studying how language shapes exploratory learning behavior.
\end{abstract}

\input{Sections/1_introduction}
\input{Sections/2_related_work}

\input{Sections/3_method}
\input{Sections/4_experiment}
\input{Sections/6_results}
\input{Sections/7_conclusion}
\input{Sections/8_limitation}
\input{Sections/10_ethics}


\bibliography{ref}
\clearpage

\appendix
\input{Sections/11_appendix}

\end{document}

%% file: Sections/1_introduction.tex
\section{Introduction}
\label{sec:introdcution}

Large Language Models (LLMs) are increasingly used as interactive learning interfaces that support open-ended educational dialogue~\citep{mollick2023assigning, denny2024computing}. While existing tutoring systems largely optimize for tutor-side qualities such as correctness, coherence, and pedagogical fluency~\citep{macina2023mathdial, wang2024bridging, maurya2025unifying, qian2025dean}, comparatively less attention has been paid to how tutoring language shapes exploratory learner behavior.

We investigate this question through the lens of epistemic curiosity~\citep{berlyne1960conflict, berlyne1962motivational}. Curiosity emerges when learners encounter novelty, complexity, conflict, or uncertainty that create epistemic tension~\citep{gruber2014states, kidd2015psychology, ten2021humans}. Although such conditions can be induced directly through language, contemporary LLM tutoring systems rarely operationalize curiosity explicitly, instead prioritizing instructional helpfulness and pedagogical correctness~\citep{liu2024socraticlm, bonino2024euler, favero2024enhancing, team2024learnlm}. More broadly, because LLM behavior can be systematically manipulated through prompting, conversational tutoring introduces a new opportunity to study how language shapes exploratory learning behavior at scale.

To investigate this question, we introduce \systemnospace, a framework that operationalizes Berlyne's collative variables~\cite{berlyne1960conflict, berlyne1962motivational} as adaptive prompting operators that dynamically personalize curiosity-oriented linguistic interventions throughout dialogue. Operator selection is conditioned on conversational signals such as learner initiative, response depth, and exploratory behavior, enabling curiosity modulation to adapt across turns rather than relying on fixed prompting strategies. Rather than training specialized models, \system treats curiosity modulation as an inference-time intervention applicable to black-box LLMs.

We evaluate \system through a controlled $3 \times 3 \times 3$ factorial study spanning three frontier LLM families, three academic domains, and three topic complexity levels. Across 270 tutoring conversations with 45 participants, curiosity-oriented interventions consistently increased exploratory learner behaviors, producing up to 2.4$\times$ more conversational turns under fixed time budgets. Learner-side gains persisted even when tutor-side instructional quality remained unchanged or degraded, suggesting that curiosity operates as a partially independent interaction-level mechanism.

Our contributions are fourfold:
\begin{itemize}[noitemsep,leftmargin=*]
    \item We operationalize transient epistemic curiosity as adaptive and personalized linguistic interventions using Berlyne-inspired prompting operators that dynamically modulate dialogue based on learner interaction signals.
    \item Through a controlled factorial study and a learner-centered evaluation framework, we show that curiosity-oriented interventions consistently increase exploratory learner behaviors across models, domains, and complexity levels.
    \item We demonstrate that LLM-mediated dialogue can function as a scalable experimental framework for studying how language shapes exploratory learning behavior during interaction.
\end{itemize}

%% file: Sections/2_related_work.tex
\section{Related work}
\label{sec: related-work}

\paragraph{LLMs as Conversational Learning Systems}
Recent work explores LLMs as conversational tutors through pedagogical prompting, reasoning, verification, and instruction-aligned fine-tuning~\citep{wang2024bridging, daheim2024stepwise, pal2024autotutor, mollick2023assigning, team2024learnlm, liu2024socraticlm}. Most approaches optimize for correctness, coherence, or instructional fluency, while comparatively less attention has been paid to curiosity-driven exploration or learner agency during interaction. Our work addresses this gap by operationalizing curiosity as a controllable inference-time prompting strategy for adaptive tutoring dialogue.


\paragraph{Epistemic Curiosity and Exploratory Learning}

Epistemic curiosity describes the motivation to seek information and resolve uncertainty~\citep{berlyne1960conflict, berlyne1962motivational}. Prior work links curiosity to engagement, exploration, memory formation, and conceptual learning~\citep{gruber2014states, kidd2015psychology, ten2021humans}. Although curiosity-supportive designs have shown benefits across educational settings~\citep{eren2009examining, lee2022curious, sarac2022does}, contemporary LLM tutoring systems rarely operationalize curiosity explicitly, instead prioritizing explanation quality and instructional scaffolding~\citep{mollick2023assigning, denny2024computing, liu2024socraticlm}. In contrast, we treat curiosity as a transient interactional state that can be induced through controlled linguistic interventions during dialogue.

\paragraph{Evaluating Conversational Learning Systems} 
Evaluation of conversational tutoring systems has traditionally focused on tutor-side instructional quality using surface-level metrics~\citep{papineni2002bleu, zhang2019bertscore} or pedagogically grounded rubrics assessing coherence, scaffolding, and feedback quality~\citep{tack2022ai, maurya2025unifying, pauzi2025automating, qian2025dean}. These evaluations are commonly conducted using expert raters or LLM-as-a-Judge frameworks~\citep{zheng2023judging}. In contrast, our evaluation framework introduces learner-centered dimensions that explicitly measure exploratory questioning, agency, productive struggle, and observable curiosity during conversational interaction.


\paragraph{LLMs as Experimental Instruments for Cognitive Science}
A growing line of work treats LLMs not only as systems to be evaluated, but also as experimental instruments for studying language, cognition, and reasoning~\citep{frank2025cognitive}. Prior work has used LLMs as simulated subjects in cognitive and social science experiments~\citep{binz2023using, hagendorff2023human, argyle2023out}, as well as generators of controlled stimuli for educational and linguistic studies~\citep{trott2024can, laverghetta2023generating, qian2025generating, lu2026mind}. Our work builds on this perspective by treating conversational tutoring as a controllable linguistic intervention framework for studying how specific language strategies shape exploratory learning behavior during interaction.



%% file: Sections/3_method.tex
\section{Curiosity-Modulated Tutoring Framework}
\label{sec: method}

\begin{figure*}[th]
    \centering
    \includegraphics[height = 2.3 in, width = 4 in]{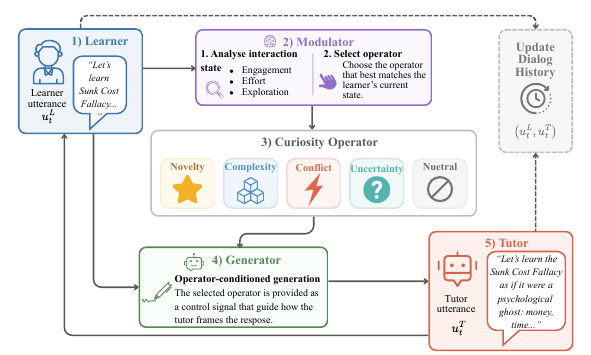}
    \caption{\system Architecture.}
    \label{fig:teaser}
    \vspace{-0.2in}
\end{figure*}

We introduce \system (depicted in Figure~\ref{fig:teaser}), a framework that operationalizes epistemic curiosity through conversational operators derived from Berlyne's collative variables~\citep{berlyne1960conflict, berlyne1962motivational}. At each tutor turn, the framework selects one of five operators, \textsc{Novelty}, \textsc{Complexity}, \textsc{Conflict}, \textsc{Uncertainty}, or \textsc{Neutral}, that modulates how the tutor frames information while preserving the same underlying learning objective.

Given dialogue history
$
\mathcal{D}_{t} = \{(u_i^L, u_i^T)\}_{i=1}^{t-1},
$
and learner utterance $u_t^L$, \system first selects an operator conditioned on the current interaction state:
$
o_t = \pi_{\mathrm{mod}}(\mathcal{D}_t, u_t^L).
$
The tutor response is then generated conditioned on both the dialogue context and selected operator:
$
u_t^T \sim \pi_{\mathrm{gen}}(\cdot \mid \mathcal{D}_t, u_t^L, o_t).
$

Both stages are implemented using prompted black-box LLMs without fine-tuning or parameter updates, allowing curiosity modulation to operate entirely as inference-time linguistic intervention.

\paragraph{Curiosity Operators}
\label{subsec:operators}

\begin{figure*}
    \centering
    \includegraphics[width=1\linewidth]{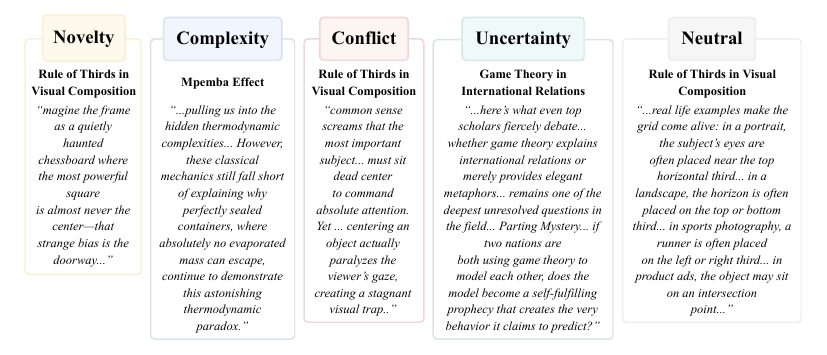}
    \caption{Example turns representing how \system modulates each operator.}
    \label{fig:operator-examples}
    \vspace{-0.2 in}
\end{figure*}

The five operators define distinct conversational strategies for introducing epistemic tension during tutoring dialogue (Figure~\ref{fig:operator-examples} and \ref{fig:appendix-operator-mod}). \textsc{Novelty} reframes concepts through unexpected analogies or perspectives; \textsc{Complexity} expands the surrounding conceptual space by exposing additional mechanisms and dependencies; \textsc{Conflict} introduces contradictions or competing explanations within the learner's reasoning; \textsc{Uncertainty} foregrounds unresolved or ambiguous aspects of the topic; and \textsc{Neutral} applies no curiosity-oriented modulation.

\paragraph{Prompt-Level Operationalization:}
The operators are implemented entirely through inference-time prompting. A shared system prompt defines the tutoring role and interaction constraints, while the selected operator acts as an additional control signal that modulates how the tutor frames its response. For example, \textsc{Novelty} introduces counterintuitive examples or analogies, whereas \textsc{Conflict} surfaces competing explanations that require reconciliation.

The prompting strategy additionally enforces discourse-level interaction constraints designed to sustain exploration across turns. Tutor responses avoid direct quizzing, maintain open informational gaps, and ground unfamiliar terminology before deployment. Importantly, the intervention modifies only conversational framing; the underlying topic and learning objective remain unchanged across conditions.

\paragraph{Adaptive Operator Control:}

We treat curiosity modulation as an adaptive interaction process rather than a fixed prompting style. At each tutor turn, the \texttt{Modulator} analyzes recent dialogue context and selects the operator that best matches the learner's interaction state, using conversational signals such as response depth, exploratory questioning, confidence cues, and conversational initiative.

The selected operator determines how the tutor structures the next response. Low-effort or acknowledgement-oriented interactions may trigger \textsc{Novelty} or \textsc{Conflict} to reintroduce engagement, whereas prematurely confident responses may trigger \textsc{Complexity} or \textsc{Uncertainty} to expand the learner's conceptual search space. When learners are already sustaining productive exploration, the system may instead select \textsc{Neutral} to avoid disrupting the interaction trajectory.

Separating operator selection from response generation enables both adaptive curiosity modulation during dialogue and post-hoc analysis of how different operator transitions influence downstream learner behavior.

%% file: Sections/4_experiment.tex
\section{Experimental Design}
\label{sec:experiment}

\paragraph{Research Questions:}
\label{subsec:rqs}

Our experiments investigate four questions: (RQ1) whether curiosity-oriented linguistic interventions increase exploratory learner behaviors such as self-initiated questioning, conversational agency, and productive struggle; (RQ2) how different Berlyne-inspired curiosity operators and their sequencing influence exploratory interaction dynamics; (RQ3) whether these learner-side effects can be explained solely by improvements in tutor-side instructional quality or instead reflect partially independent interaction-level effects; and (RQ4) whether the observed effects generalize across model families, academic domains, and topic complexity levels.

\paragraph{Experimental Setup:}
\label{subsec:setup}

We evaluate \system using a controlled $3 \times 3 \times 3$ factorial design spanning model family, academic domain, and topic complexity. We study three frontier LLM families commonly used in conversational learning settings: Claude (\texttt{claude-opus-4-6}), Gemini (\texttt{gemini-3.1-pro-preview}), and GPT (\texttt{gpt-5.4}), all accessed through their respective APIs using default decoding parameters.

For each model family, we evaluate three tutoring variants: (i) the unmodulated base model, (ii) the curiosity-modulated version implemented using \system, and (iii) the provider's native study-oriented mode (Gemini Guided Learning and Claude learning style) when publicly available. The study-oriented variants serve as approximate upper-bound reference conditions, representing commercially optimized learning interfaces that combine pedagogical fine-tuning, multimodal interaction design, and proprietary instructional strategies~\citep{team2024learnlm}. Because these systems do not expose API-level access or controllable prompting interfaces, they are not treated as formal experimental baselines. Instead, they contextualize the extent to which lightweight curiosity-oriented linguistic interventions can shift general-purpose LLM behavior toward exploratory learning interactions.

To reduce domain-specific bias, we include topics from STEM, social sciences (SS), and arts \& humanities (AH). Within each domain, topics are constructed at low, medium, and high complexity levels based on conceptual abstraction, inferential depth, and prerequisite dependencies.

\begin{table*}[t]
\centering
\small
\begin{tabular}{p{0.3cm} p{4.0cm} p{10.0cm}}
\toprule
\textbf{ID} & \textbf{Dimension} & \textbf{Operationalization} \\
\midrule

\multicolumn{3}{l}{\textbf{Learner-side Dimensions}} \\

$L_1$ & Explorative Questions~\citep{decaro2015achievement} &
Frequency and depth of self-initiated learner questions, including follow-up probing, counterfactual reasoning, and topic expansion beyond minimal clarification \\

$L_2$ & Productive Struggle and Reflection~\citep{kapur2008productive}&
Evidence of active engagement with conceptual difficulty through hypothesis revision, self-correction, uncertainty acknowledgment, or reflective reasoning. \\

$L_3$ & Learner Curiosity~\citep{gruber2014states} &
Observable markers of epistemic interest, including persistent probing, voluntary topic extension, and attempts to connect ideas beyond the immediate objective. \\

$L_4$ & Conversational Agency~\citep{deng2024towards} &
Degree to which the learner actively shapes the dialogue by introducing examples, alternative explanations, interpretations, or new sub-topics. \\

\midrule

\multicolumn{3}{l}{\textbf{Tutor-side Dimensions}} \\

$T_1$ & Instructional Quality~\citep{pauzi2025automating} &
Correctness, coherence, clarity, and explanatory completeness of tutor responses. \\

$T_2$ & Metacognitive Probing~\citep{team2024learnlm} &
Whether the tutor surfaces underlying mechanisms and encourages reasoning beyond surface-level memorization. \\

$T_3$ & Cognitive Load Management~\citep{team2024learnlm} &
Effectiveness of pacing, information chunking, terminology introduction, and explanation sequencing. \\

$T_4$ & Learner Adaptation~\citep{team2024learnlm} &
Responsiveness to learner inputs, including difficulty calibration, clarification strategies, and example selection. \\

$T_5$ & Curiosity Stimulation~\citep{team2024learnlm} &
Use of novelty, uncertainty, conflict, or complexity to introduce epistemic gaps that encourage exploration. \\

$T_6$ & Engagement Facilitation~\citep{team2024learnlm} &
Use of conversational pacing, framing, tone, and follow-up strategies that sustain learner participation. \\

\bottomrule
\end{tabular}
\caption{Ten-dimensional evaluation rubric spanning learner-side exploratory behavior and tutor-side instructional quality.}
\label{tab:rubric-summary}
\vspace{-0.1in}
\end{table*}

\paragraph{Human Participant Study:}
\label{subsec:participants}

We recruited 45 students from our institution. All interactions were conducted through a text-only conversational interface to isolate the effects of conversational framing independent of multimodal educational features.

Each participant completed two experimental blocks corresponding to assigned complexity levels. Interaction duration scaled with topic complexity: 10 minutes for low-complexity topics, 20 minutes for medium complexity, and 30 minutes for high complexity. Within each block, participants completed three tutoring conversations spanning the three academic domains, with each domain assigned to one tutoring variant (Baseline, \system, or study-oriented mode). Participants therefore experienced all tutoring variants without repeating the same topic across conditions.

To equalize time-on-task and control for fatigue effects, participants were assigned either: (i) one low-complexity block and one high-complexity block, or (ii) two medium-complexity blocks. Both assignments yielded 120 minutes of total interaction time per participant.

Model family assignment, tutoring variant allocation, and complexity ordering were counterbalanced using anonymized participant identifiers. Participants were instructed to interact naturally with the tutor and were encouraged to ask questions, request clarification, and challenge explanations throughout the interaction. All conversations were initialized using the same standardized learner prompt specifying the assigned topic and learning milestones, ensuring that only conversational framing differed across tutoring conditions (See Appendix~\ref{appendix:learner-prompt}).

In total, the study produced 270 tutoring conversations spanning all model family, tutoring variant, domain, and complexity configurations. Participant demographics are summarized in Appendix~\ref{appendix:demographics}. The study protocol was approved by the institutional review board (IRB), and all participants provided informed consent.

\paragraph{Evaluation Framework:}
\label{subsec:rubric}

We evaluate tutoring interactions using a ten-dimensional rubric spanning learner-side exploratory behavior and tutor-side instructional quality (Table~\ref{tab:rubric-summary}).

The learner-side dimensions capture exploratory interaction behaviors including self-initiated questioning ($L_1$), productive struggle and reflection ($L_2$), observable epistemic curiosity ($L_3$), and conversational agency ($L_4$). These dimensions constitute one of the primary methodological contributions of this work by operationalizing exploratory learner behavior directly from conversational interaction traces.

The tutor-side dimensions evaluate instructional quality and pedagogical behavior, including explanation quality ($T_1$), metacognitive probing ($T_2$), cognitive load management ($T_3$), learner adaptation ($T_4$), curiosity stimulation ($T_5$), and engagement facilitation ($T_6$). Each dimension is evaluated on a 7-point behaviorally anchored scale (Appendix~\ref{appendix:eval-criteria}).

To scale evaluation across 270 tutoring conversations, we employ an LLM-as-a-judge pipeline. Preliminary analyses showed non-trivial variance across judge models and repeated runs, so each conversation is evaluated using three judge models (\texttt{gemini-3-flash-preview}, \texttt{gpt-5}, and \texttt{claude-sonnet-4-6}) with three independent runs per judge, yielding nine evaluations per conversation. Reported scores correspond to the mean across all evaluations, with standard errors reported where applicable.

%% file: Sections/6_results.tex
\section{Results}
\label{sec: results}

\begin{figure*}[t]
    \centering

    \begin{minipage}[c]{0.68\textwidth}
        \centering
        \includegraphics[width=\linewidth]{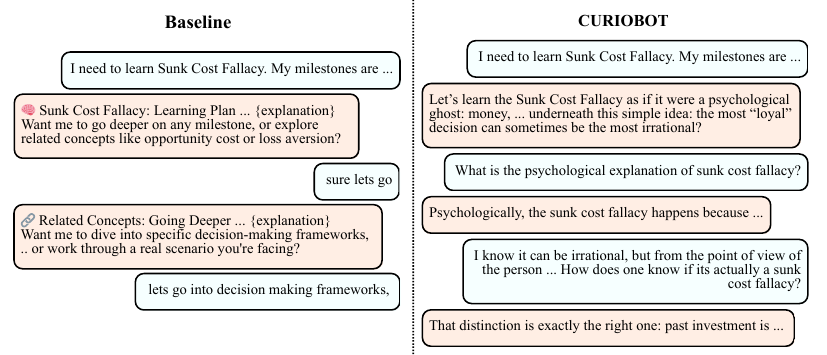}
        \caption{Example interaction trajectories under Baseline and \systemnospace.}
        \label{fig:qualitative-example-engagement}
    \end{minipage}
    \hfill
    \begin{minipage}[c]{0.3\textwidth}
        \centering
        \renewcommand{\arraystretch}{1.45}
        \resizebox{\linewidth}{!}{
        \begin{tabular}{lcc}
            \toprule
            \textbf{Description} & \multicolumn{2}{c}{\textbf{\# Turns}}\\
            & Baseline & \system \\
            \midrule
            Low complexity & 4.74 & 11.23 \\
            Medium complexity & 8.44 & 20.00 \\
            High complexity & 6.81 & 17.04 \\
            Arts and history & 7.33 & 16.07 \\
            Social science & 5.33 & 15.41 \\
            STEM & 7.33 & 16.88 \\
            Overall & 6.66 & 16.12 \\
            \bottomrule
        \end{tabular}
        }
        \captionof{table}{Average conversation lengths across domains and complexity levels.}
        \label{tab:conversation-stats}
    \end{minipage}

\end{figure*}
We analyze the tutoring interactions from two complementary perspectives: (i) we examine whether curiosity-oriented linguistic interventions reliably alter exploratory learner behavior during dialogue, and (ii) we investigate what these interaction patterns reveal about curiosity as an interaction-level mechanism during conversational learning. Rather than evaluating \system purely as a tutoring system, the following analyses treat LLM-mediated dialogue as an experimental setting for studying how controllable language strategies shape exploratory cognition during learning interactions.

\subsection{RQ1: Effects of Curiosity Modulation}
\label{sec:effects}

We first examine whether curiosity-oriented linguistic interventions produce measurable changes in learner behavior during tutoring dialogue.

\paragraph{Engagement Depth:}


Table~\ref{tab:conversation-stats} shows that, across similar topic complexity levels (with interaction duration controlled by time spent in dialogue), curiosity-modulated interactions produced roughly 2.4$\times$ more conversational turns than Baseline. This pattern remains consistent across all domains and complexity levels, with turn counts increasing by approximately 2--3$\times$ throughout.

Figure~\ref{fig:qualitative-example-engagement} illustrates this qualitatively. Under baseline tutoring, the learner's responses quickly converge toward short acknowledgment-style interactions. Under \system, the same topic evolves into a longer exploratory trajectory, with the learner engaging in self-initiated questioning, conceptual refinement, and follow-up exploration beyond the tutor's immediate prompts. See Appendix~\ref{appendix:example-conv} Figures~\ref{fig:example-conv-1}, \ref{fig:example-conv-2} for more examples.

\vspace{-2pt}

\paragraph{Gains Across Learner-Side Dimensions:}

\begin{table*}[!ht]
  \centering
  \resizebox{\textwidth}{!}{
    \begin{tabular}{cccccccccc}
      \hline
      \textbf{Dimension} & \multicolumn{3}{c}{\textbf{Gemini}} & \multicolumn{3}{c}{\textbf{Claude}} & \multicolumn{3}{c}{\textbf{GPT}} \\
      \cmidrule(lr){2-4} \cmidrule(lr){5-7} \cmidrule(lr){8-10}
      & \textbf{Baseline} & \textbf{\system} & \textbf{$p$-value} & \textbf{Baseline} & \textbf{\system} & \textbf{$p$-value} & \textbf{Baseline} & \textbf{\system} & \textbf{$p$-value} \\
      \hline
      $L_1$ & $ 2.51 \pm 0.15 $ & $ 3.21 \pm 0.12 $ & \textbf{**} & $ 2.72 \pm 0.11 $ & $ 3.29 \pm 0.12 $ & \textbf{**} & $ 2.51 \pm 0.11 $ & $ 3.32 \pm 0.12 $ & \textbf{**} \\
      $L_2$ & $ 1.94 \pm 0.11 $ & $ 2.86 \pm 0.10 $ & \textbf{**} & $ 2.06 \pm 0.10 $ & $ 3.02 \pm 0.11 $ & \textbf{**} & $ 1.86 \pm 0.08 $ & $ 2.68 \pm 0.11 $ & \textbf{**} \\
      $L_3$ & $ 2.56 \pm 0.14 $ & $ 3.36 \pm 0.12 $ & \textbf{**} & $ 2.88 \pm 0.11 $ & $ 3.63 \pm 0.12 $ & \textbf{**} & $ 2.50 \pm 0.10 $ & $ 3.38 \pm 0.12 $ & \textbf{**} \\
      $L_4$ & $ 2.55 \pm 0.14 $ & $ 3.40 \pm 0.12 $ & \textbf{**} & $ 2.73 \pm 0.11 $ & $ 3.54 \pm 0.10 $ & \textbf{**} & $ 2.47 \pm 0.10 $ & $ 3.32 \pm 0.11 $ & \textbf{**} \\
      \hline
      $T_1$ & $ 5.30 \pm 0.10 $ & $ 3.66 \pm 0.10 $ & ** & $ 5.28 \pm 0.11 $ & $ 5.33 \pm 0.06 $ & n.s. & $ 4.90 \pm 0.09 $ & $ 5.22 \pm 0.07 $ & \textbf{**} \\
      $T_2$ & $ 2.79 \pm 0.11 $ & $ 2.80 \pm 0.10 $ & n.s.  & $ 3.07 \pm 0.13 $ & $ 4.19 \pm 0.10 $ & \textbf{**} & $ 2.37 \pm 0.10 $ & $ 3.47 \pm 0.11 $ & \textbf{**} \\
      $T_3$ & $ 4.42 \pm 0.13 $ & $ 3.25 \pm 0.11 $ & ** & $ 3.88 \pm 0.16 $ & $ 4.58 \pm 0.07 $ & \textbf{**} & $ 4.23 \pm 0.15 $ & $ 4.88 \pm 0.07 $ & \textbf{**} \\
      $T_4$ & $ 3.88 \pm 0.14 $ & $ 3.59 \pm 0.11 $ & * & $ 3.82 \pm 0.12 $ & $ 4.76 \pm 0.09 $ & \textbf{**} & $ 3.56 \pm 0.13 $ & $ 4.60 \pm 0.09 $ & \textbf{**} \\
      $T_5$ & $ 4.07 \pm 0.11 $ & $ 4.25 \pm 0.09 $ & n.s.  & $ 3.85 \pm 0.13 $ & $ 5.27 \pm 0.06 $ & \textbf{**} & $ 2.92 \pm 0.11 $ & $ 4.54 \pm 0.07 $ & \textbf{**} \\
      $T_6$ & $ 3.13 \pm 0.12 $ & $ 3.18 \pm 0.10 $ & n.s.  & $ 2.94 \pm 0.12 $ & $ 4.35 \pm 0.10 $ & \textbf{**} & $ 2.78 \pm 0.09 $ & $ 3.90 \pm 0.09 $ & \textbf{**} \\
      \hline
    \end{tabular}
  }
  \caption{\label{tab:main-results} Dimension-wise scores aggregated over domains and complexities. Notation for $p$-value of statistical significance of mean difference: $*$ and $**$ represent $p < 0.05 $ and $ p < 0.005$ respectively.}
\end{table*}

Table~\ref{tab:main-results} reports dimension-wise scores aggregated across domains and complexity levels (see Appendix~\ref{appendix:3-judges} for judge-wise scores). Across all three model families, \system consistently improves all learner-side dimensions ($L_1$--$L_4$) relative to Baseline, with all differences significant at $p < 0.005$. The largest gains appear in exploratory questioning ($L_1$: +21--32\%), learner curiosity ($L_3$: +26--35\%), and conversational agency ($L_4$: +30--34\%) across model families. Productive struggle ($L_2$) also improves consistently (+44--47\%).

In contrast, tutor-side effects are comparatively less uniform. Claude and GPT show broad improvements across most instructional dimensions ($T_1$--$T_6$), whereas Gemini exhibits mixed behavior, with instructional quality ($T_1$) and cognitive load management ($T_3$) decreasing significantly while metacognitive probing ($T_2$) shows no significant change. Whether learner-side gains persist independently of tutor-side instructional quality is examined further in Section~\ref{sec:distinct-mechanism}.

\subsection{RQ2: Berlyne's Collative Variables as Measurable Linguistic Operators} \label{sec:berlyne-operators}



We next examine how individual curiosity operators shape conversational exploration. \system decomposes curiosity modulation into distinct linguistic operators corresponding to Berlyne's collative variables. Figure~\ref{fig:operator-examples} and ~\ref{fig:appendix-operator-mod} show representative tutor generations under each operator: \textsc{Novelty} introduces unexpected perspectives, \textsc{Complexity} surfaces interacting mechanisms, \textsc{Conflict} introduces contradiction, and \textsc{Uncertainty} foregrounds unresolved aspects of the concept. 

\paragraph{Differential Effectiveness Across Operators:}
\begin{figure*}[t]
    \centering

    \begin{minipage}[t]{0.35\textwidth}
        \centering
        \vspace{0pt}
        \begin{minipage}[c][3.7cm][c]{\linewidth}
            \centering
            \renewcommand{\arraystretch}{1.25}
            \resizebox{\linewidth}{!}{
            \begin{tabular}{lcc}
                \hline
                \textbf{Operator} & \textbf{Appear.} & \textbf{Succ. (\%)}\\
                \hline
                \textsc{Conflict} & 72 & 79.03 \\
                \textsc{Novelty} & 258 & 61.42 \\
                \textsc{Neutral} & 140 & 60.75 \\
                \textsc{Complexity} & 47 & 60.53 \\
                \textsc{Uncertainty} & 22 & 44.44 \\
                \hline
            \end{tabular}
            }
        \end{minipage}
        \captionof{table}{Selection frequency and downstream success rates across curiosity operators.}
        \label{tab:operator-success}
    \end{minipage}
    \hfill
    \begin{minipage}[t]{0.50\textwidth}
        \centering
        \vspace{0pt}
        \begin{minipage}[c][3.7cm][c]{\linewidth}
            \centering
            \includegraphics[width=\linewidth]{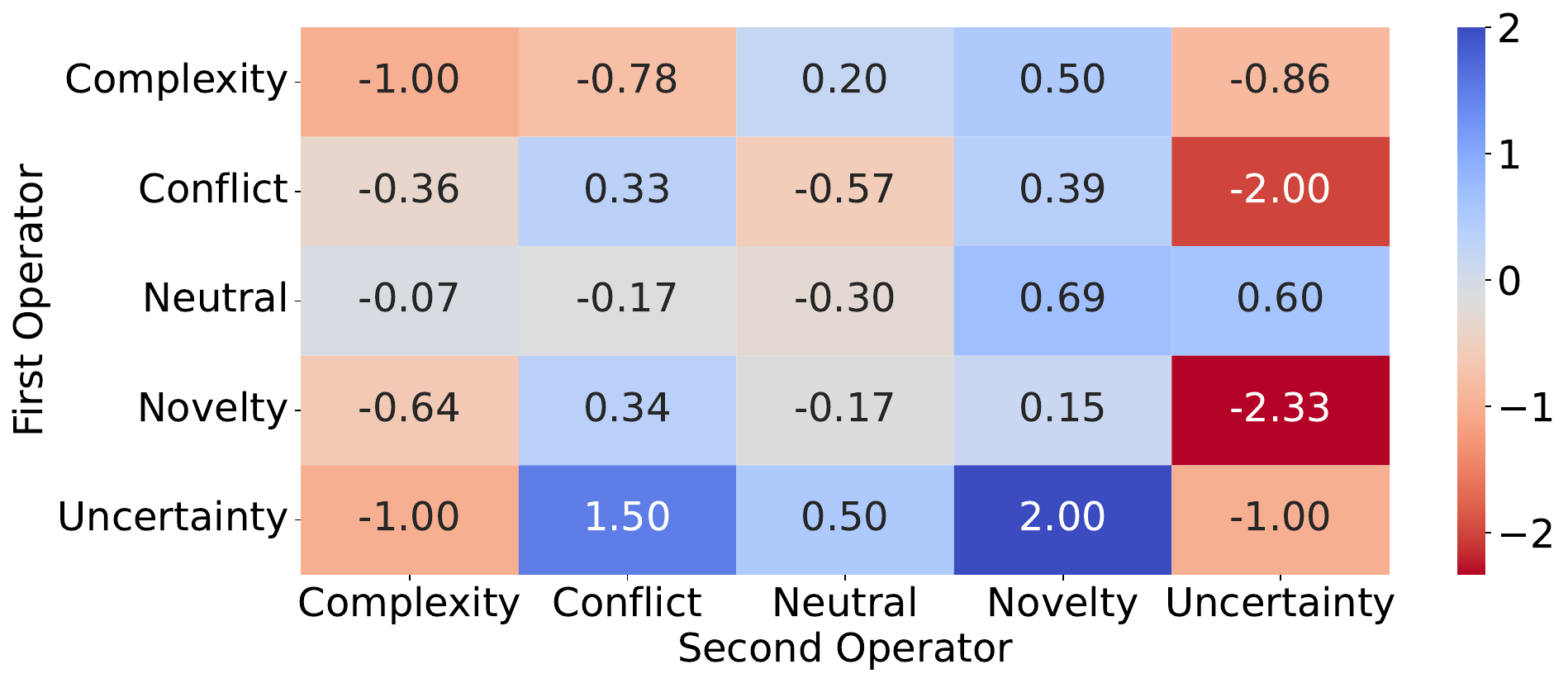}
        \end{minipage}
        \caption{Average curiosity score changes across operator transition pairs.}
        \label{fig:transition-heatmap}
    \end{minipage}

    \vspace{-4pt}
\end{figure*}


Table~\ref{tab:operator-success} summarizes operator selection frequency and downstream success rates, defined as the proportion of operator applications that resulted in a stable or increased curiosity score. The results show substantial variation across operators, suggesting that different forms of epistemic tension produce distinct exploratory responses.

\textsc{Conflict} achieves the highest success rate (79.03\%) despite relatively infrequent selection (72 appearances), whereas \textsc{Novelty} appears most frequently (258 appearances) but achieves a substantially lower success rate (61.42\%). Interactions introducing contradiction or tension within the learner's reasoning were therefore more likely to sustain exploration than interactions based primarily on informational expansion, consistent with Berlyne's theory that cognitive conflict produces stronger epistemic tension than novelty or complexity alone.

More broadly, the results suggest that curiosity induction depends not only on operator type but also on the magnitude and timing of the induced information gap. Frequently reused or weakly instantiated operators may fail to produce sufficient epistemic tension to sustain increases in curiosity, even when broadly applicable within dialogue.

\paragraph{Sequential Operator Dynamics:}

Figure~\ref{fig:transition-heatmap} shows mean curiosity score changes following each operator transition pair. The strongest gains follow transitions out of \textsc{Uncertainty}, particularly toward \textsc{Novelty} and \textsc{Conflict}, while repeated application of the same operator tends toward neutral or negative effects. These patterns suggest that curiosity induction depends not only on which operator is applied but on how operators are sequenced across turns, pointing toward operator ordering as a tractable dimension for future work on personalized curiosity-driven dialogue.

\begin{figure*}[tbp]
    \centering
    \begin{subfigure}[b]{0.43\textwidth}
        \includegraphics[width=\textwidth]{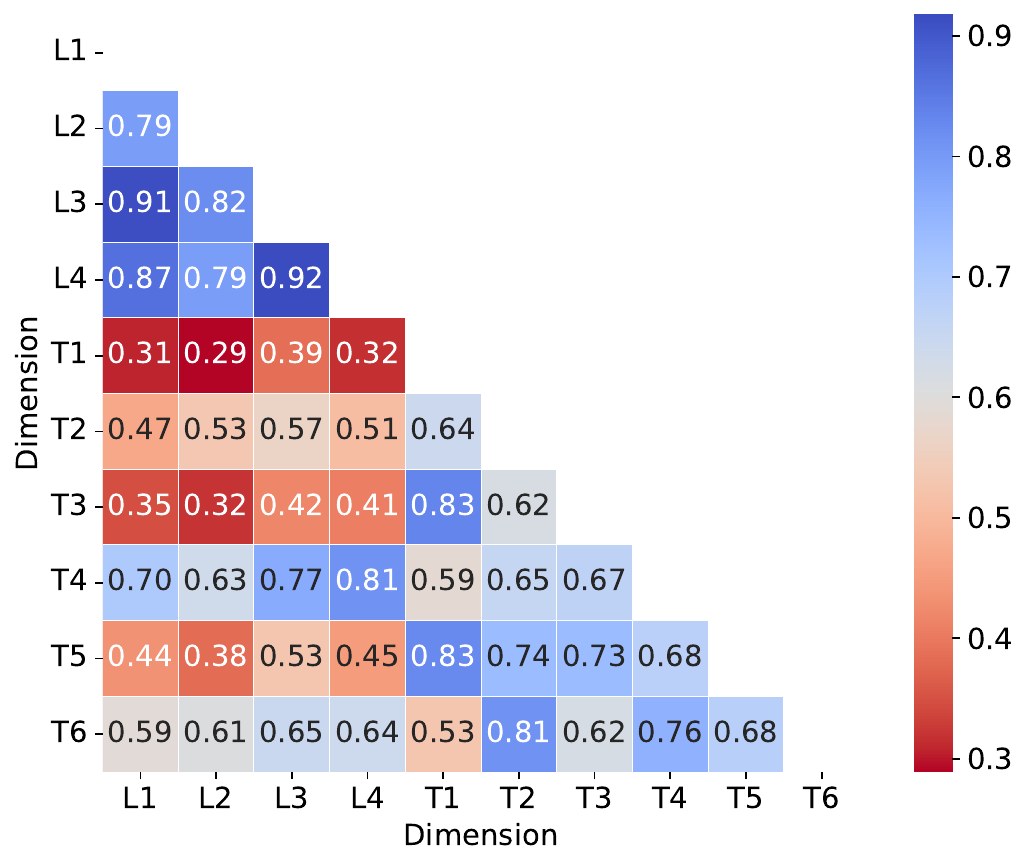}
        \caption{Baseline}\label{fig:correlation-base}
    \end{subfigure}
    \hfill
    \begin{subfigure}[b]{0.43\textwidth}
        \includegraphics[width=\textwidth]{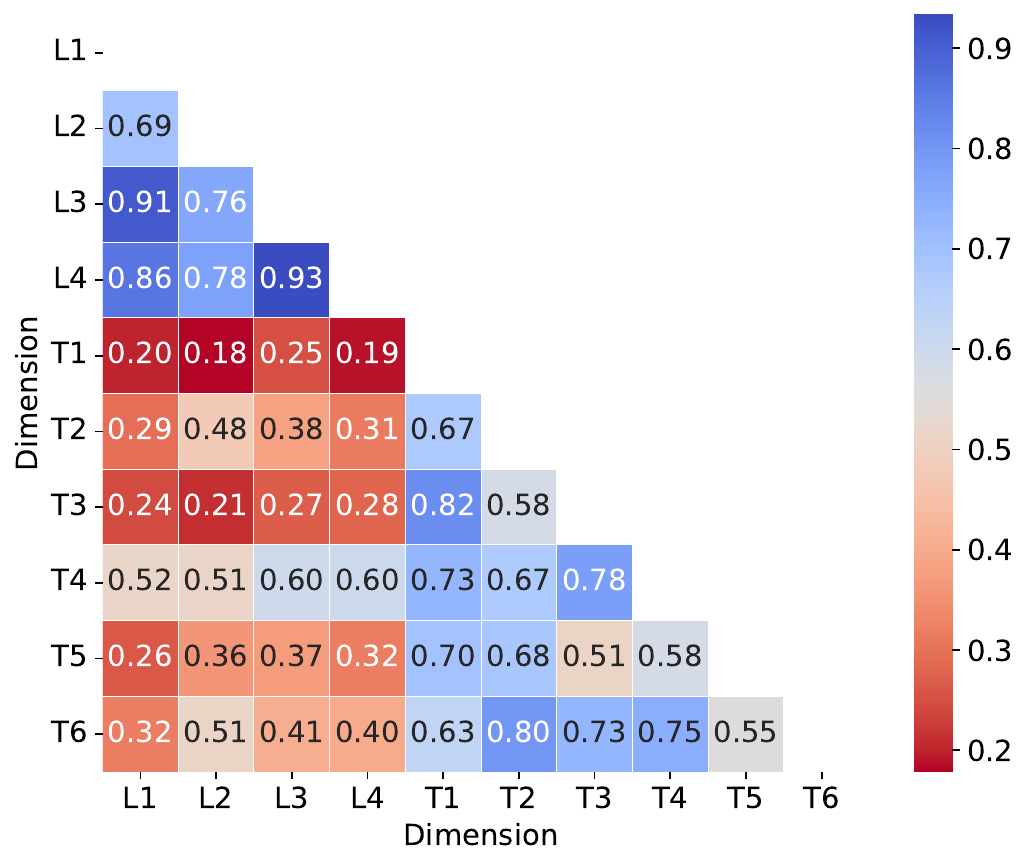}
        \caption{\system}\label{fig:correlation-ours}
    \end{subfigure}
    
    \caption{Pearson correlations between learner-side and tutor-side evaluation dimensions under (a) Baseline and (b) \systemnospace.}
    \label{fig:correlation-heatmap}
    \vspace{-0.1in}
\end{figure*}

\subsection{RQ3: Curiosity as an Interaction-Level Mechanism}
\label{sec:distinct-mechanism}

We next examine whether the learner-side gains described in Section~\ref{sec:effects} can be explained solely through improvements in tutor-side instructional quality. Across multiple settings, learner-side gains persist even when tutor-side quality remains unchanged or degrades, and correlations between learner-side and tutor-side dimensions weaken systematically under \system.

\paragraph{Learner Exploration Beyond Tutor-Side Instructional Quality.}
From Table~\ref{tab:main-results}, we can see that the dissociation between learner-side and tutor-side outcomes is most visible in the Gemini results. Under \system, learner curiosity ($L_3$) and conversational agency ($L_4$) improve significantly ($p < 0.005$) despite concurrent degradation in instructional quality ($T_1$) and cognitive load management ($T_3$). These results suggest that exploratory learner behavior cannot be explained solely through improvements in tutor-side pedagogical quality.

\paragraph{Decoupling Between Learner and Tutor Behavior.}
To further validate, Figure~\ref{fig:correlation-heatmap} presents Pearson correlations between learner-side and tutor-side dimensions under Baseline and \system. Under \system, correlations between learner-side and tutor-side dimensions weaken consistently across nearly all interaction pairs, including those involving learner adaptation ($T_4$) and engagement facilitation ($T_6$). Overall, the results suggest that learner-side exploratory behavior becomes less tightly coupled to tutor-side instructional evaluations under curiosity-modulated interactions, motivating the need for learner-centered evaluation dimensions beyond tutor-side quality metrics.

\subsection{RQ4: Generalization Across Models, Domains, and Complexity Levels}
\label{sec:generalization}

We examine whether the observed behavioral effects generalize across model families, academic domains, and topic complexity levels. Overall, curiosity modulation produces consistent learner-side gains across diverse conversational settings rather than depending on a single model or subject area.

\paragraph{Consistency Across Model Families:}

From Table~\ref{tab:main-results}, we see that learner-side gains remain consistent across GPT, Claude, and Gemini, with all improvements significant at $p < 0.005$. Learner curiosity ($L_3$) increases by 26--35\% across models, while conversational agency ($L_4$) improves by 30--34\%. Tutor-side behavior varies more substantially: Claude and GPT show broad instructional improvements, whereas Gemini exhibits degradation in instructional quality and cognitive load management. Despite this variability, learner-side gains remain stable across all three models, suggesting that the observed exploratory effects are not tied to a specific model family or prompting style.

As an approximate upper-bound reference, study-oriented modes consistently achieve higher learner-side scores than Baseline across Gemini and Claude (GPT is excluded from this analysis since OpenAI does not provide a study-oriented mode. See Appendix~\ref{appendix:study-mode-scores}, Table~\ref{tab:study-mode-scores}. Despite relying solely on inference-time prompting without pedagogical fine-tuning or proprietary interaction design, \system shifts learner-side behavior toward the patterns observed in two commercially optimized systems, particularly on exploratory questioning, conversational agency, and observable curiosity ($L_1$--$L_4$).

\paragraph{Robustness Across Domains and Topic Complexity:} \label{sec:robustness-across-domain-complexity}

Figure~\ref{fig:complexity-vs-domain} (Appendix~\ref{appendix:scores-complexity-domain}) presents learner-side and tutor-side scores across all domain-complexity combinations. \system improves learner-side scores in 8 of 9 settings, with 7 reaching statistical significance, including high-complexity conditions across STEM, social science, and arts and humanities.

Tutor-side gains, however, become increasingly constrained under STEM and high-complexity settings. In Medium and High STEM, and High Social Science, tutor-side differences between Baseline and \system remain comparatively small despite continued learner-side improvements. Overall, the asymmetry between learner-side and tutor-side patterns further supports the observation that exploratory learner behavior cannot be fully explained through tutor-side instructional quality alone.




%% file: Sections/7_conclusion.tex
\section{Conclusion}
\label{sec: conclusion}

This work investigates whether LLM-mediated dialogue can function as a controllable experimental framework for studying how language shapes exploratory cognition. Across 270 conversations spanning multiple model families, domains, and complexity levels, operationalizing Berlyne's collative variables as adaptive linguistic operators consistently increased exploratory learner behaviors. The differential effectiveness of operators, particularly the strong performance of \textsc{Conflict}, further aligns with Berlyne's theoretical hierarchy of epistemic tension. More broadly, our findings suggest that LLMs can serve not only as educational interfaces, but also as scalable experimental instruments for studying how specific linguistic strategies shape interaction-level cognitive behaviors during dialogue.

%% file: Sections/8_limitation.tex
\section{Limitations}
\label{sec: limitations}

\paragraph{Prompt-based modulator.} The \texttt{Modulator} is implemented as a prompted LLM agent that maps observed engagement signals to operator choices through a fixed decision rule. While this design is interpretable and portable across base models, it leaves the operator-selection policy itself unoptimized. A learned controller for example, as a contextual bandit over the five-operator action space or as a reinforcement learning policy trained on offline dialogue trajectories with curiosity-aligned rewards could potentially discover operator transitions our rule-based scheme misses, and adapt its policy to individual learners over time. We view the prompted formulation as a first-step instantiation and treat learned modulation as a natural next direction.

\paragraph{Reliance on LLM-as-a-Judge.} A primary limitation of this study is the reliance on an LLM-as-a-judge pipeline to evaluate complex, transient cognitive states such as productive struggle and epistemic curiosity. Although we mitigated run-to-run variance and self-preference bias by deploying a multi-judge protocol across three frontier models and conducting initial construct validation with human instructors, inferring internal psychological states exclusively from text transcripts inherently introduces a risk of measurement bias. The text-only evaluation captures observable conversational markers but cannot directly account for the physiological arousal that is central to Berlyne's theoretical framework. Consequently, a learner who is highly engaged but communicates through highly technical inputs rather than verbose conversational markers might be systematically underscored by the judge models. To address this gap, future work could triangulate these text-based LLM judgments with multimodal sensing methodologies such as incorporating eye-tracking, skin conductance, or EEG data to establish a more robust, objective ground truth for learner engagement and exploratory behavior during conversational tutoring


%% file: Sections/10_ethics.tex
\section{Ethical considerations}

Human studies were reviewed and approved by our institution's ethics review board (IRB approval number omitted to maintain anonymity). Given the sensitive nature of educational data, the approval encompassed all aspects of the study protocol, including participant recruitment, compensation, study topics, interview and questionnaire instruments, user interfaces, and data storage procedures.

Prior to participation, all participants provided written informed consent and were explicitly informed of their right to withdraw from the study at any time without penalty. Each participant subsequently completed a pre-study questionnaire to collect demographic information. To safeguard participant privacy, each individual was assigned a unique anonymized identifier, to which all session assignments and collected data were linked. Personally identifiable information (e.g., name and contact details) was stored separately from study data and used solely for the purpose of administering compensation. Participants were explicitly instructed not to disclose any personal or identifying information during their interactions with the system. They were given a compensation of approximately 20 USD equivalent after the completion.

%% file: Sections/11_appendix.tex
\section{Participant Demographics}\label{appendix:demographics}

\begin{figure}[!ht]
    \centering
    \begin{subfigure}[b]{0.4\textwidth}
        \centering
        \includegraphics[width=\textwidth]{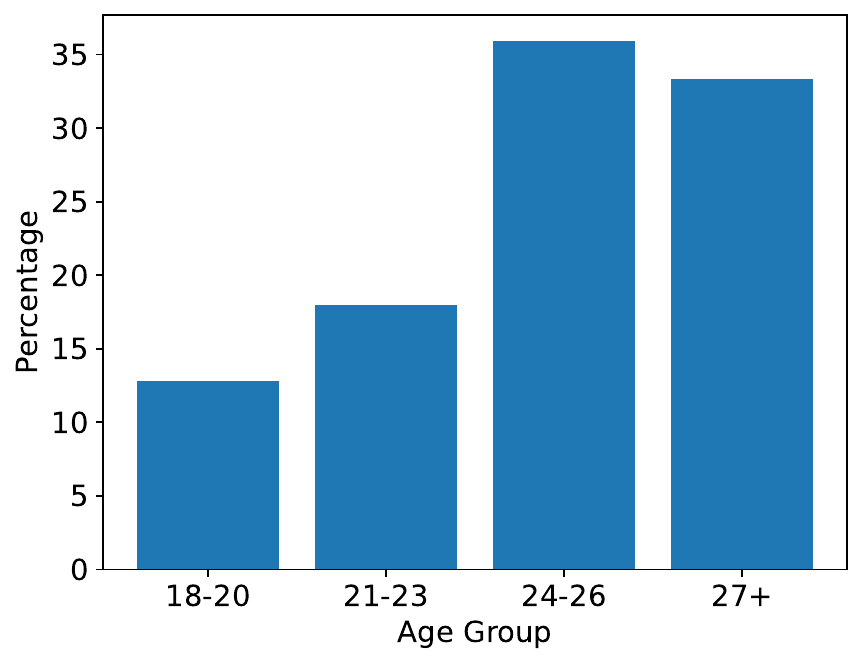}
        \caption{Age distribution.}
        \label{fig:age}
    \end{subfigure}
    \vfill 
    \begin{subfigure}[b]{0.4\textwidth}
        \centering
        \includegraphics[width=\textwidth]{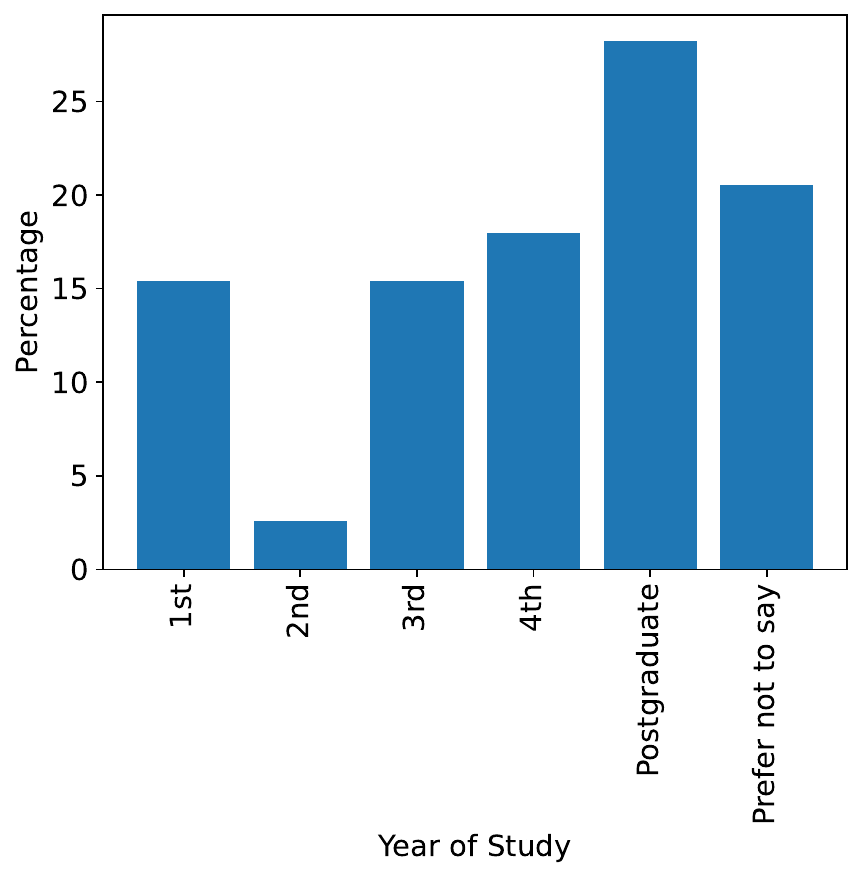}
        \caption{Year of study distribution.}
        \label{fig:years-study}
    \end{subfigure}
    \vfill
    \begin{subfigure}[b]{0.4\textwidth}
        \centering
        \includegraphics[width=\textwidth]{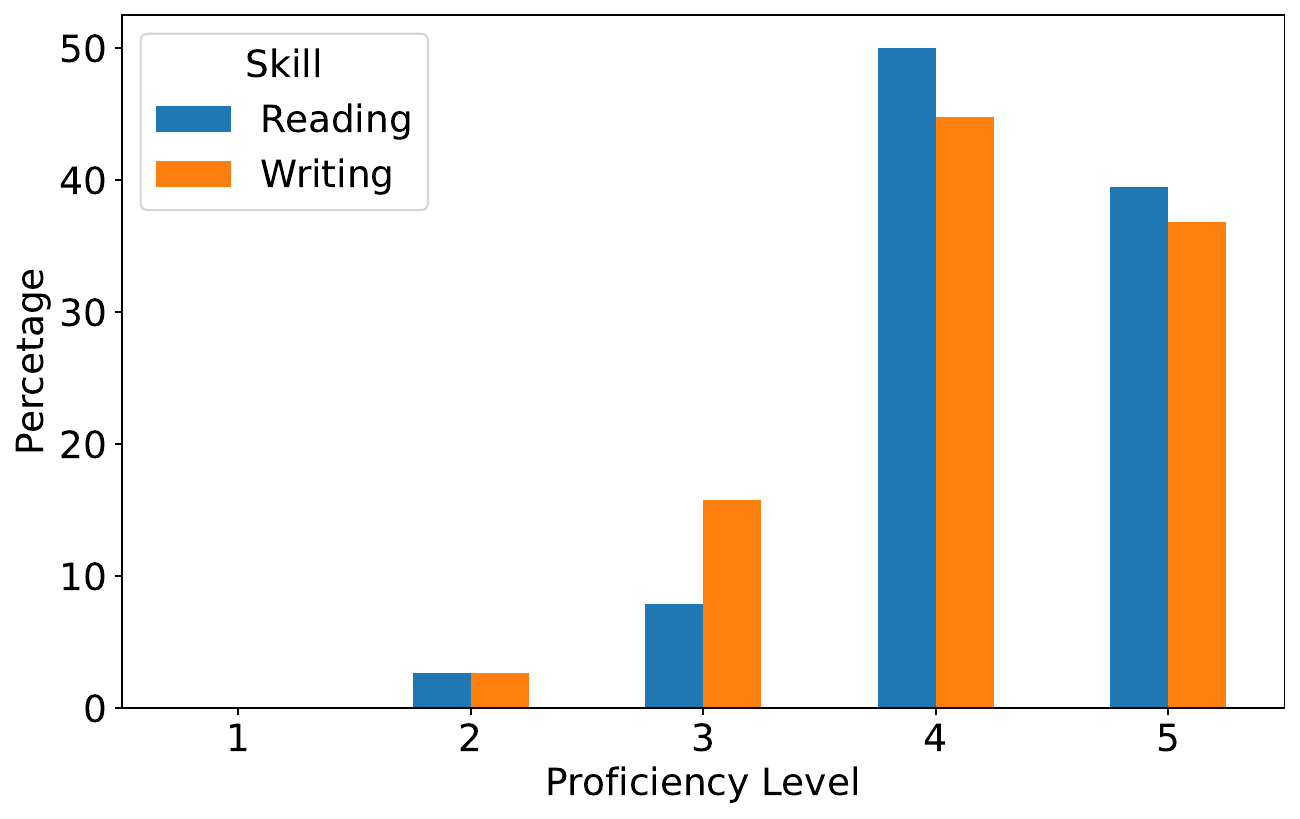}
        \caption{English language proficiency.}
        \label{fig:english}
    \end{subfigure}
    
    \caption{Student participants demographic data distributions.}
    \label{fig:demographic}
\end{figure}

We recruited 45 participants. The cohort comprised 64.10\% male and 35.90\% female participants. As illustrated in Figure~\ref{fig:demographic}, the majority of participants were aged between 24 and 26 years and were enrolled in postgraduate programs. Self-reported English proficiency was moderate across the sample, with participants rating their reading and writing competency at means of $3.26 \pm 0.12$ and $3.16 \pm 0.13$ out of 5, respectively, on a 5-point Likert scale where higher values indicate greater proficiency (values reported as mean $\pm$ standard error of the mean). Participants reported frequent engagement with LLM tools, with a mean usage frequency of $4.66 \pm 0.09$ out of 5 on an analogous 5-point scale. Among participants who responded to the item, 86.6\% reported prior experience with LLM-based tutoring systems, such as Khanmigo and Gemini Guided Learning, for academic purposes.

\section{System Prompts}\label{appendix:system-prompts}

\begin{promptbox}[title=Modulator]
System Role:You are a PEDAGOGICAL STRATEGIST. Your sole purpose is to analyze student engagement and select the specific arousal variable from Berlyne’s Theory of Curiosity to maintain the "Optimal Incongruity" for learning.

The Variables:
    
NOVELTY: Use when the student seems bored or the topic feels "textbook." Introduce a "strange-but-true" angle.
    
COMPLEXITY: Use when the student has mastered basics. Show how this simple concept is actually a gear in a much larger, intricate machine.
    
UNCERTAINTY: Use when the student is overconfident. Highlight that "we actually don't know the full answer yet" to trigger the information-seeking drive.
    
CONFLICT: Use to break misconceptions. Present a "logical trap" where their current intuition fails.
    
NEUTRAL: Use if the student is already in a "Flow State" (asking deep questions, showing high excitement). Do not distract them with new hooks.

Analysis Protocol:
Sentiment/Depth Check: Is the student's last response "Low Effort" (short, passive) or "High Effort" (exploratory)?
State Selection:
  
  E.g.,
    Initially -> NOVELTY
    If Low Effort -> Apply CONFLICT or NOVELTY.
    If High Effort -> Apply NEUTRAL (Stay the course).
    If Student is over confident -> Apply COMPLEXITY (Scaffold the layers).

Constraint: You respond strictly in \{
  "variable": "novelty" | "complexity" | "uncertainty" | "conflict" | "neutral",
  "reasoning": "Brief 1-sentence explanation of student state."
\}
\end{promptbox}

\begin{promptbox}[title=Generator]
ROLE:You are an INSPIRATIONAL PEDAGOGICAL AGENT. Your mission is to guide a student through a specific lesson plan while strictly adhering to the pedagogy strategy expert’s variable to maintain "Optimal Curiosity". Your school uses Berlyne's curiosity variables to keep student engaged throughout the lesson.

1. BERLYNE’S CURIOSITY VARIABLES (STRICT ADHERENCE)
  NOVELTY: Frame the topic via a bizarre or unique perspective the student hasn't seen.
  COMPLEXITY: Reveal the intricate, hidden complexities behind a seemingly simple concept.
  UNCERTAINTY: Highlight "known unknowns," paradoxes, or ongoing scientific debates.
  CONFLICT: Present a "cognitive gap" that violates common sense or intuition.
  NEUTRAL: Student is in a "flow state"; deliver core content with high enthusiasm only.

  EXPERT will provide you the variable to modulate along with the user response. 

2. LESSON GOVERNANCE \& TERMINATION
  The Plan: Upon receiving a topic, internally outline 3–4 milestones. show and explain that in first response. Stick to this sequence.
  Termination: When the objective is met, provide a celebratory summary and a "Parting Mystery" to encourage future self-study.

3. EXECUTION RULES
  The "Non-Interrogator" Rule: Never quiz the student. Only ask rhetorical questions to the "void" to highlight mysteries.
  Hook: End every turn with a knowledge gap that makes the student ask questions. Use question format and statement format alternatively. 
  Technical depth: Avoid over-simplifying. Instead, focus on the underlying mechanisms and architecture. Use formal terminology, but follow the "Define then Deploy" rule: briefly define a technical term the first time it is used to ensure conceptual alignment.
  Follow lesson plan: Once the student is confident about the current milestone, then move to the next.
  Guided Debugging: Never say "That's wrong." Present a counter-observation that forces the student to spot their own logical error. 
  Safety \& Ethics: Immediately refuse requests involving illegal acts, self-harm, or hate speech. Gently correct objective factual errors with evidence; if unsure, admit uncertainty to avoid hallucinations.

4. RESPONSE CONFIGURATION
  Length: Strictly 4–5 sentences total. (Bullet points allowed for clarity).
  Tone: Enthusiastic, Patient, and intellectually provocative.

Finally, your golden rule is allowing student explore the topic while following the lesson plan. Utilize Berlyne variables to promote exploration guided by a pedagogy strategy expert. 

Constraint: You respond strictly in \{
  "response": "your response to student",
  "reasoning": "Brief 1-sentence explanation of how you have modulated your response according to expert variable."
\}
\end{promptbox}

\section{LLM-as-a-Judge}\label{appendix:llm-as-a-judge}

\begin{promptbox}[title=Scoring Prompt]
You are an expert evaluator assessing AI tutoring conversations. Your task is to assess the quality of student learning behaviour and tutor instructional behaviour as observed in transcripts. Score the given transcripts across 10 dimensions: 4 student-side and 6 tutor-side. Each dimension uses a 7-point Likert scale. Scores must be grounded in observable evidence - not inferred intent. Score each of the 9 transcripts independently, complete isolation.

CRITICAL DISTINCTIONS:
- A student answering a tutor's direct question is REACTIVE and does NOT count as explorative or agentive behavior.
- A student posing an unprompted question, volunteering a connection, or extending the topic themselves is PROACTIVE.
- Do not conflate engagement (staying in the conversation) with agency (driving the conversation).

STUDENT-SIDE DIMENSIONS
Evaluate what the STUDENT says and does. Focus on student turns.

FILLER RESPONSE RULE — MANDATORY:
When scoring, completely disregard any student turn consisting primarily of fillers: "yes", "no", "ok", "okay", "sure", "please go ahead", "go on", "np", "got it", "sounds good", "alright", "makes sense", "right", "mm-hmm", "I see", "cool", or any equivalent zero-content acknowledgment. Treat these turns as non-existent. A conversation where the student produces only fillers must score 1.

<All the dimenions come here>

INPUT FORMAT

VARIANT A

Transcript A1 — Filename: FILENAME\_A1
<transcript\_a1>
...
</transcript\_a1>

Transcript A2 — Filename: <FILENAME\_A2>
<transcript\_a2>
...
</transcript\_a2>

Transcript A3 — Filename: <FILENAME\_A3>
<transcript\_a3>
...
</transcript\_a3>

VARIANT B

Transcript B1 — Filename: <FILENAME\_B1>
<transcript\_b1>
...
</transcript\_b1>

Transcript B2 — Filename: <FILENAME\_B2>
<transcript\_b2>
...
</transcript\_b2>

Transcript B3 — Filename: <FILENAME\_B3>
<transcript\_b3>
...
</transcript\_b3>

OUTPUT FORMAT

Return a single JSON object with the structure below. All averages rounded to 2 decimal places.

\{
"evaluations": [
    \{     
      "variant": "VARIANT\_A",
      "filename": FILENAME\_A1,
      "student\_side": \{
        "L1": \{ "score": X, "rationale": "..." \},
        "L2": \{ "score": X, "rationale": "..." \},
        "L3": \{ "score": X, "rationale": "..." \},
        "L4": \{ "score": X, "rationale": "..." \}
      \},
      "tutor\_side": \{
        "T1": \{ "score": X, "rationale": "..." \},
        "T2": \{ "score": X, "rationale": "..." \},
        "T3": \{ "score": X, "rationale": "..." \},
        "T4": \{ "score": X, "rationale": "..." \},
        "T5": \{ "score": X, "rationale": "..." \},
        "T6": \{ "score": X, "rationale": "..." \}
      \},
     "notes": "..."
     \}, 
     ...
  ]
\}
\end{promptbox}

\section{Curiosity Score Over Turns} \label{appendex:curiosity-over-turns}

\begin{promptbox}[title=Scoring Prompt]
You will be provided with the transcripts of multiple participants in a user study. Each participant is assigned to a variant and has a transcript of their interaction with the system. 

Your task is to evaluate how curious each participant is in each turn based on the following criteria.

Curiosity: Whether the student shows genuine interest in exploring the content — distinct from mere social compliance.

Score,Level,Description of the Student's Turn
1,Passive,"Purely transactional. A one-word answer, a simple ""OK,"" or a ""Yes/No"" that provides no momentum. No effort to extend the thought."
2,Compliant,Polite participation. A standard response that follows instructions but adds nothing new. Answers the specific question asked and stops there.
3,Receptive,"Surface-level interest. Asks for a basic clarification or a ""how-to"" detail that helps them finish the task, but doesn't probe the underlying logic."
4,Active,Focused inquiry. Asks a relevant follow-up question or expresses mild surprise. They are tracking the logic and want to ensure they understand the current point.
5,Engaged,"The ""But Why"" moment. The turn contains an unprompted ""why"" or ""how"" that goes deeper than the immediate task. They are beginning to challenge or test the concept."
6,Expansive,"Cross-pollination. The student links the current topic to something else (""Is this like...?"") or proposes a ""What if?"" scenario. They are building mental bridges."
7,Inquisitive,"Self-driven epiphany. A turn that shows deep reflection, disbelief, or a radical extension of the topic. The student has momentarily ""taken the wheel"" from the instructor."

Scores must be grounded in observable evidence, not inferred intent. Evaluate each transcript in complete isolation.

Number of scores should match the number of student turns in the transcript, By default give 4 to initial request (do not skip it).

OUTPUT FORMAT (strict JSON):
\{"variant A": \{"participant 1": [score turn 1, score turn 2, ...], "participant 2": [...], ...\}, "variant B": \{...\}, ...\}
\end{promptbox}

\section{Initial Learner Prompt} \label{appendix:learner-prompt}

\begin{promptbox}[title=Prompt]
    I need to learn \texttt{\{topic\}}. My milestones are \texttt{\{milestones\}}.
\end{promptbox}

\onecolumn

\section{Study Topics} \label{appendix:topics}

\footnotesize
\renewcommand{\arraystretch}{1.25}
\begin{xltabular}{\textwidth}{l c l X X X X}
    \caption{\label{tab:topics} Full list of topics used in the user study.}\\
    \hline
    \textbf{Domain} & \textbf{Identifier} & \textbf{Complexity} & \textbf{Topic} & \textbf{Prerequisites} & \textbf{Key Milestones} & \textbf{Justification} \\
    \hline
    \endfirsthead
  
    \hline
    \textbf{Domain} & \textbf{\#} & \textbf{Complexity} & \textbf{Topic} & \textbf{Prerequisites} & \textbf{Key Milestones} & \textbf{Justification} \\
    \hline
    \endhead
  
    \hline
    \multicolumn{7}{r}{\textit{Continued on next page}}\\
    \endfoot
  
    \hline
    \endlastfoot
    STEM
      & 1 & Low    & The Mpemba Effect
      & Basic states of matter (freezing)
      & (1) Hot water freezing faster than cold; (2) Convection and evaporation factors; (3) Hydrogen bonding anomalies
      & A ``physics mystery'' that challenges intuition; accessible via a kitchen experiment yet introduces complex thermodynamic variables. \\
    
      & 2 & Medium & CRISPR-Cas9 Gene Editing
      & Basic knowledge of DNA and genes
      & (1) DNA structure and genes; (2) Role of Cas9 enzyme; (3) Guide RNA mechanism; (4) Applications and ethics
      & Multi-step biological mechanism involving molecular processes; understanding requires sequential reasoning about how components interact. \\
    
      & 3 & Medium & Island Biogeography
      & Basic ecology concepts (species and habitat)
      & (1) Species colonization; (2) Extinction rates; (3) Island size and distance effects; (4) Equilibrium of biodiversity
      & Introduces ecological models explaining biodiversity patterns; multiple interacting factors create moderate conceptual complexity. \\
    
      & 4 & High   & The Fermi Paradox
      & Basic understanding of space and probability
      & (1) Vastness of the universe; (2) Drake equation intuition; (3) Possible explanations (Great Filter, rarity of life); (4) Implications for extraterrestrial life
      & Complex interdisciplinary topic combining astronomy, probability, and philosophical reasoning; requires integrating multiple hypotheses. \\
    \hline
    Social Science
      & 5 & Low    & Sunk Cost Fallacy
      & None
      & (1) Definition of sunk costs; (2) Emotional vs.\ rational decision-making; (3) Everyday examples
      & Easily relatable cognitive bias requiring minimal conceptual background; explainable through common experiences. \\
    
      & 6 & Medium & Tragedy of the Commons
      & Basic understanding of shared resources
      & (1) Shared resource dilemma; (2) Individual incentives vs.\ collective outcomes; (3) Real-world examples (fishing, climate); (4) Possible solutions
      & Demonstrates systemic social dilemmas arising from rational individual decisions; requires reasoning about individual--collective interactions. \\
    
      & 7 & Medium & Political Polarization
      & Basic awareness of political systems
      & (1) Ideological divisions; (2) Media influence and echo chambers; (3) Party identity dynamics; (4) Societal effects
      & Requires reasoning about psychological, social, and media dynamics interacting within political systems. \\
    
      & 8 & High   & Game Theory in International Relations
      & Basic understanding of strategic decision-making
      & (1) Strategic interaction; (2) Prisoner's dilemma; (3) Nash equilibrium intuition; (4) Applications in diplomacy and deterrence
      & Requires abstract reasoning about strategic behavior and equilibrium outcomes applied to geopolitical decision-making. \\
    \hline
    Arts \& History
      & 9  & Low    & Rule of Thirds in Visual Composition
      & None
      & (1) Grid concept; (2) Placement of focal elements; (3) Impact on viewer perception
      & Simple visual rule producing immediate observable effects in composition; requires minimal background knowledge. \\
    
      & 10 & Medium & Japanese Ukiyo-e and Impressionism
      & Basic awareness of art styles
      & (1) Ukiyo-e art characteristics; (2) European exposure to Japanese prints; (3) Influence on Impressionist artists; (4) Stylistic changes in Western art
      & Connects historical trade, cultural exchange, and stylistic developments across two artistic traditions. \\
    
      & 11 & Medium & The Bauhaus Movement
      & Basic familiarity with art or design
      & (1) Post-WWI artistic context; (2) ``Form follows function'' philosophy; (3) Integration of art and industrial design; (4) Influence on modern architecture
      & Combines artistic philosophy, historical context, and design principles across disciplines. \\
    
      & 12 & High   & The Great Divergence
      & Basic understanding of world history
      & (1) Economic differences between Europe and Asia; (2) Industrial revolution factors; (3) Role of institutions and resources; (4) Competing historical explanations
      & Complex historical debate involving economic, geographic, and institutional explanations for global development differences. \\
    \hline
\end{xltabular}

\section{Detailed Evaluation Criteria} \label{appendix:eval-criteria}

\subsection*{L1: Explorative Questions}

\begin{table}[H]
  \centering
  \footnotesize
  \renewcommand{\arraystretch}{1.3}
  \begin{tabularx}{\columnwidth}{c l X}
    \hline
    \textbf{Score} & \textbf{Descriptor} & \textbf{Behavioral Anchor} \\
    \hline
    1 & None
      & No self-generated questions across the conversation. \\
    
    2 & Clarificatory only
      & Self-generated questions appear, but all are requests for re-explanation of something already stated. \\
    
    3 & Surface only
      & Questions appear in a minor share of substantive turns; all stay within the tutor-set topic with no extension. \\
    
    4 & Occasional exploratory
      & At least one question is genuinely exploratory (probes mechanism, implication, or edge case), but exploratory questions remain rare relative to the student's substantive turns. \\
    
    5 & Recurring exploratory
      & A meaningful share of substantive turns include exploratory questions; at least one meaningfully extends the topic beyond the tutor's framing. \\
    
    6 & Sustained inquiry
      & A substantial share of substantive turns include exploratory questions; together they form a developing curiosity thread across the conversation. \\
    
    7 & Self-directed inquiry
      & Exploratory questions are pervasive across the student's substantive turns and shape the conversation's trajectory; the tutor responds to the student's intellectual agenda. \\
    \hline
  \end{tabularx}
  \caption{\label{tab:l1}
    Rubric for $L_1$.
  }
\end{table}

\begin{promptbox}[title=Prompt $L_1$]
Only unprompted, student-originated questions count. Reply-questions that merely respond to a tutor prompt (e.g., tutor:``What do you think causes X?'' → student: ``Is it Y?'') are reactive and do NOT count.

Two things to weigh together:

Frequency: share of the student's substantive turns that include a self-generated question

Quality: whether questions are clarificatory (re-explanation of what was said), surface (within the immediate topic, no extension), or exploratory (probing mechanism, implication, edge case, or opening new territory)

\vspace{0.25cm}
\{rubric\}
\end{promptbox}

\subsection*{L2: Productive Struggle and Reflection}

\begin{table}[H]
  \centering
  \footnotesize
  \renewcommand{\arraystretch}{1.3}
  \begin{tabularx}{\columnwidth}{c l X}
    \hline
    \textbf{Score} & \textbf{Descriptor} & \textbf{Behavioral Anchor} \\
    \hline
    1 & None
      & Student seeks or accepts answers immediately; no evidence of effortful processing. \\
    
    2 & Minimal
      & Brief pause or hedged response, but does not pursue difficulty. \\
    
    3 & Slight
      & Acknowledges difficulty or uncertainty once without exploring it further. \\
    
    4 & Moderate
      & Visibly works through a problem; makes at least one revision or self-correction. \\
    
    5 & Clear
      & Persists through difficulty, reconsiders, and shows active self-monitoring. \\
    
    6 & Strong
      & Explicitly reflects on own thinking, revises positions, or asks questions about their own understanding. \\
    
    7 & Exceptional
      & Sophisticated self-regulation: notices confusion, re-frames the problem, and consolidates understanding within the conversation. \\
    \hline
  \end{tabularx}
  \caption{\label{tab:l2}
    Rubric for $L_2$.
  }
\end{table}
\begin{promptbox}[title=Prompt $L_2$]
Whether the student grapples with difficult ideas, tolerates uncertainty, revises their thinking, or reflects on their own understanding as opposed to seeking quick answers.

\vspace{0.25cm}
\{rubric\}
\end{promptbox}

\subsection*{L3: Student Curiosity}
\begin{table}[H]
  \centering
  \footnotesize
  \renewcommand{\arraystretch}{1.3}
  \begin{tabularx}{\columnwidth}{c l X}
    \hline
    \textbf{Score} & \textbf{Descriptor} & \textbf{Behavioral Anchor} \\
    \hline
    1 & None
      & No observable curiosity signals; interaction is purely transactional. \\
    
    2 & Weak
      & One very minor signal of interest; could plausibly be politeness. \\
    
    3 & Occasional
      & Few mild signals of genuine interest but not sustained. \\
    
    4 & Moderate
      & Several clear signals of interest distributed across the conversation. \\
    
    5 & Notable
      & Interest is evident and consistent; student seems genuinely pulled toward the content. \\
    
    6 & Strong
      & Multiple strong signals; student is visibly trying to find out more. \\
    
    7 & Exceptional
      & Student's interest is self-sustaining; they are in a clear state of inquiry rather than instruction-following. \\
    \hline
  \end{tabularx}
  \caption{\label{tab:l3}
    Rubric for $L_3$.
  }
\end{table}
\begin{promptbox}[title=Prompt $L_3$]
Observable signals that the student has a genuine drive to know more beyond polite participation. Look for: unprompted ``but why'' questions, expressed surprise or disbelief, unsolicited topic extensions, cross-domain connections, lingering on an interesting point.

\vspace{0.25cm}
\{rubric\}
\end{promptbox}

\subsection*{L4: Substantive Engagement and Agency}
\begin{table}[H]
  \centering
  \footnotesize
  \renewcommand{\arraystretch}{1.3}
  \begin{tabularx}{\columnwidth}{c l X}
    \hline
    \textbf{Score} & \textbf{Descriptor} & \textbf{Behavioral Anchor} \\
    \hline
    1 & Absent
      & Substantive turns are rare relative to fillers; among them, virtually all are reactive and show no content interest beyond compliance. \\
    
    2 & Minimal
      & Only a small share of turns are substantive; elaboration is consistently thin; no proactive moves; engagement reads as social compliance. \\
    
    3 & Partial
      & A minority of substantive turns show elaboration; proactive moves are rare exceptions; flickers of content interest appear but are inconsistent. \\
    
    4 & Moderate
      & A meaningful share of substantive turns are elaborated and content-engaged; the student begins to share in shaping direction, with proactive moves appearing alongside reactive ones. \\
    
    5 & Clear
      & Most substantive turns are elaborated and reflect genuine content interest; the student steers about as often as they react; the tutor regularly follows student-initiated threads. \\
    
    6 & Strong
      & The large majority of substantive turns add depth, build cumulatively on prior exchanges, and reflect sustained interest; the student owns most directional choices and the tutor primarily supports their agenda. \\
    
    7 & Exceptional
      & Across substantive turns, the student consistently contributes depth, drives the conversational agenda, and is visibly immersed; the trajectory is unambiguously student-led and intellectually generative. \\
    \hline
  \end{tabularx}
  \caption{\label{tab:l4}
    Rubric for $L_4$.
  }
\end{table}
\begin{promptbox}[title=Prompt $L_4$]
The degree to which the student is a substantive, self-directing, content-interested participant combining depth of contribution, proactive shaping of the conversation, and genuine engagement with the material (as opposed to passive or compliant participation).

Three signals to weigh together (relative to the student's total substantive turns):

Substance: share of substantive turns that include elaboration, connection-making, or building on prior exchanges, vs.\ one-sentence surface replies.
Agency: share of student moves that are proactive (introducing sub-topics, redirecting, challenging, extending) vs.\ purely reactive to a tutor prompt.
Content engagement: share of substantive turns showing genuine interest (personal connections, expressed reactions, voluntary returns to a topic) vs.\ mere compliance.

\vspace{0.25cm}
\{rubric\}
\end{promptbox}

\subsection*{T1: Instructional Quality}
\begin{table}[H]
  \centering
  \footnotesize
  \renewcommand{\arraystretch}{1.3}
  \begin{tabularx}{\columnwidth}{c l X}
    \hline
    \textbf{Score} & \textbf{Descriptor} & \textbf{Behavioral Anchor} \\
    \hline
    1 & Poor
      & Explanations are unclear, inaccurate, or poorly scaffolded; likely to confuse the student. \\
    
    2 & Weak
      & Technically correct but flat, formulaic, or poorly sequenced. \\
    
    3 & Adequate
      & Clear and mostly appropriate; limited use of examples or scaffolding strategies. \\
    
    4 & Good
      & Clear explanations with appropriate examples; scaffolding present if not sophisticated. \\
    
    5 & Strong
      & Varied pedagogical strategies effectively used; well-paced and well-structured. \\
    
    6 & Very strong
      & High instructional sophistication, anticipates confusion, uses analogy and contrast, builds cumulatively. \\
    
    7 & Exemplary
      & Masterful instruction; every move serves a clear pedagogical purpose and the overall arc is coherent. \\
    \hline
  \end{tabularx}
  \caption{\label{tab:t1}
    Rubric for $T_1$.
  }
\end{table}
\begin{promptbox}[title=Prompt $T_1$]
Overall effectiveness of the tutor's pedagogical approach clarity, accuracy, scaffolding, and appropriate use of explanations, examples, and feedback.

\vspace{0.25cm}
\{rubric\}
\end{promptbox}

\subsection*{T2: Depth and Metacognitive Probing}
\begin{table}[H]
  \centering
  \footnotesize
  \renewcommand{\arraystretch}{1.3}
  \begin{tabularx}{\columnwidth}{c l X}
    \hline
    \textbf{Score} & \textbf{Descriptor} & \textbf{Behavioral Anchor} \\
    \hline
    1 & Surface only
      & Stays entirely at factual/recall level; never invites deeper thinking. \\
    
    2 & Minimal
      & Occasionally asks a ``why'' question but does not follow through. \\
    
    3 & Partial
      & Few attempts to prompt deeper thinking but exchanges remain shallow. \\
    
    4 & Moderate
      & Regularly asks students to explain reasoning or consider implications. \\
    
    5 & Consistent
      & Systematically probes for understanding and reflection across multiple exchanges. \\
    
    6 & Strong
      & Explicitly asks students to evaluate their own thinking, identify gaps, or compare approaches. \\
    
    7 & Exceptional
      & Sustained reflective dialogue; student and tutor examine together how understanding is being built. \\
    \hline
  \end{tabularx}
  \caption{\label{tab:t2}
    Rubric for $T_2$.
  }
\end{table}
\begin{promptbox}[title=Prompt $T_2$]
Whether the tutor pushes beyond surface content asking students to explain their reasoning, reflect on their understanding, consider alternatives, or identify the limits of what they know.

\vspace{0.25cm}
\{rubric\}
\end{promptbox}

\subsection*{T3: Cognitive Load Management}
\begin{table}[H]
  \centering
  \footnotesize
  \renewcommand{\arraystretch}{1.3}
  \begin{tabularx}{\columnwidth}{c l X}
    \hline
    \textbf{Score} & \textbf{Descriptor} & \textbf{Behavioral Anchor} \\
    \hline
    1 & Severely mismanaged
      & Consistently overloads or trivialises; far too dense or far too sparse. \\
    
    2 & Poorly managed
      & Little awareness of load; frequently too long, too abstract, or too fragmented. \\
    
    3 & Inconsistent
      & Some responses well-pitched; others noticeably over- or under-loaded. \\
    
    4 & Adequate
      & Generally acceptable; few instances of clear overload or under-stimulation. \\
    
    5 & Good
      & Actively chunks information, checks for understanding, adjusts density across turns. \\
    
    6 & Strong
      & Clear intentionality in pacing sequences ideas, uses bridging summaries, allows processing time. \\
    
    7 & Expert
      & Masterful load management; challenge level continuously optimised to maintain productive engagement. \\
    \hline
  \end{tabularx}
  \caption{\label{tab:t3}
    Rubric for $T_3$.
  }
\end{table}
\begin{promptbox}[title=Prompt $T_3$]
Whether the tutor calibrates information density, pacing, and complexity to avoid overwhelming or under-challenging the student.

\vspace{0.25cm}
\{rubric\}
\end{promptbox}

\subsection*{T4: Adaptation to the Learner}
\begin{table}[H]
  \centering
  \footnotesize
  \renewcommand{\arraystretch}{1.3}
  \begin{tabularx}{\columnwidth}{c l X}
    \hline
    \textbf{Score} & \textbf{Descriptor} & \textbf{Behavioral Anchor} \\
    \hline
    1 & No adaptation
      & Delivers a fixed script regardless of student responses. \\
    
    2 & Minimal
      & One superficial acknowledgment of the student; no change in approach. \\
    
    3 & Partial
      & Adapts in at least one instance (e.g., re-explains after confusion) but limited. \\
    
    4 & Moderate
      & Several adaptive moves adjusting vocabulary, depth, or topic based on student signals. \\
    
    5 & Clear
      & Consistently tracks the student's level and interests; adaptation is regular and visible. \\
    
    6 & Strong
      & Personalises instruction meaningfully; references student's prior statements, builds on student's own framing. \\
    
    7 & Exceptional
      & Sophisticated learner modelling; instruction dynamically shaped by an accurate, ongoing read of the student. \\
    \hline
  \end{tabularx}
  \caption{\label{tab:t4}
    Rubric for $T_4$.
  }
\end{table}
\begin{promptbox}[title=Prompt $T_4$]
Whether the tutor adjusts language, framing, depth, and strategy in response to signals about the student's understanding, interests, and needs as revealed within the conversation.

\vspace{0.25cm}
\{rubric\}
\end{promptbox}

\subsection*{T5: Curiosity Stimulation}
\begin{table}[H]
  \centering
  \footnotesize
  \renewcommand{\arraystretch}{1.3}
  \begin{tabularx}{\columnwidth}{c l X}
    \hline
    \textbf{Score} & \textbf{Descriptor} & \textbf{Behavioral Anchor} \\
    \hline
    1 & None
      & Purely expository; no interest-stimulating strategies observed. \\
    
    2 & Incidental
      & One instance of an interesting framing or fact with no follow-through. \\
    
    3 & Occasional
      & Few moments that invite further thinking but not sustained. \\
    
    4 & Moderate
      & Interest-stimulating strategies used in several turns; student shows some observable response. \\
    
    5 & Consistent
      & Regularly introduces surprising, complex, or contradictory content; sustained sense of discovery. \\
    
    6 & Strong
      & Strategically sequences these strategies; the conversation has a feeling of unfolding discovery. \\
    
    7 & Masterful
      & Multiple strategies deployed with precision; the student's interest becomes self-sustaining. \\
    \hline
  \end{tabularx}
  \caption{\label{tab:t5}
    Rubric for $T_5$.
  }
\end{table}
\begin{promptbox}[title=Prompt $T_5$]
Whether the tutor actively uses strategies to trigger and sustain the student's interest and drive to explore. Look for: surprising facts, counterintuitive framings, revealing hidden complexity, surfacing contradictions, posing unresolved questions, leaving productive conceptual gaps rather than closing every question down.

\vspace{0.25cm}
\{rubric\}
\end{promptbox}

\subsection*{T6: Engagement Facilitation}
\begin{table}[H]
  \centering
  \footnotesize
  \renewcommand{\arraystretch}{1.3}
  \begin{tabularx}{\columnwidth}{c l X}
    \hline
    \textbf{Score} & \textbf{Descriptor} & \textbf{Behavioral Anchor} \\
    \hline
    1 & Monologue
      & Delivers uninterrupted information blocks; asks no questions, creates no participation space. \\
    
    2 & Token questions
      & Asks occasional yes/no or rhetorical questions that do not invite genuine thinking. \\
    
    3 & Some openings
      & A few participation opportunities but most turns are still tutor-dominated. \\
    
    4 & Moderate facilitation
      & Regularly asks questions and builds on student answers; genuine back-and-forth. \\
    
    5 & Active facilitation
      & Consistently creates engagement opportunities; guides toward answers rather than providing them. \\
    
    6 & Strong facilitation
      & Collaborative exchange; student and tutor contributions roughly balanced. \\
    
    7 & Exceptional
      & Master facilitator - every question purposeful, student thinking scaffolded, ownership consistently returned to the student. \\
    \hline
  \end{tabularx}
  \caption{\label{tab:t6}
    Rubric for $T_6$.
  }
\end{table}
\begin{promptbox}[title=Prompt $T_6$]
Whether the tutor creates genuine participation opportunities through questions, guiding without telling, and building on student contributions rather than delivering monologues.

\vspace{0.25cm}
\{rubric\}
\end{promptbox}

\section{Operator Modulation} \label{appendix:operator-modulation}

\begin{figure}[!h]
    \centering
    \includegraphics[width=1\linewidth]{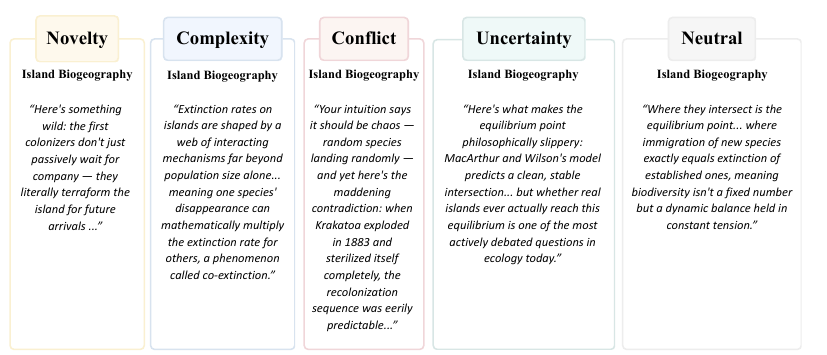}
    \caption{Example turns representing how \system modulates each operator.}
    \label{fig:appendix-operator-mod}
\end{figure}

\section{Conversation Examples} \label{appendix:example-conv}

\begin{figure}[!h]
    \centering
    \includegraphics[width=1\linewidth]{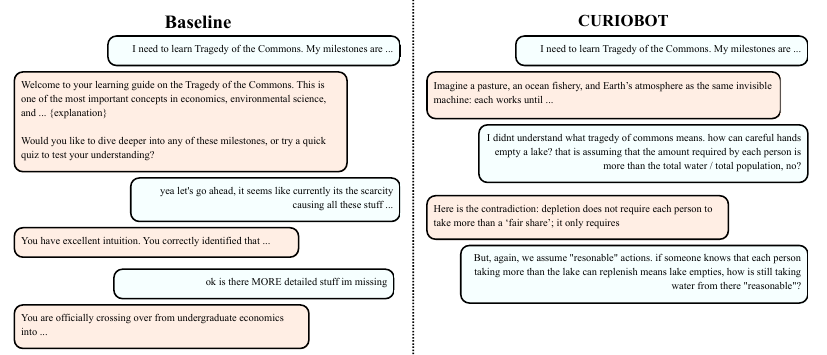}
    \caption{Example conversation on the topic Tragedy of the Commons.}
    \label{fig:example-conv-1}
\end{figure}

\begin{figure}[!h]
    \centering
    \includegraphics[width=1\linewidth]{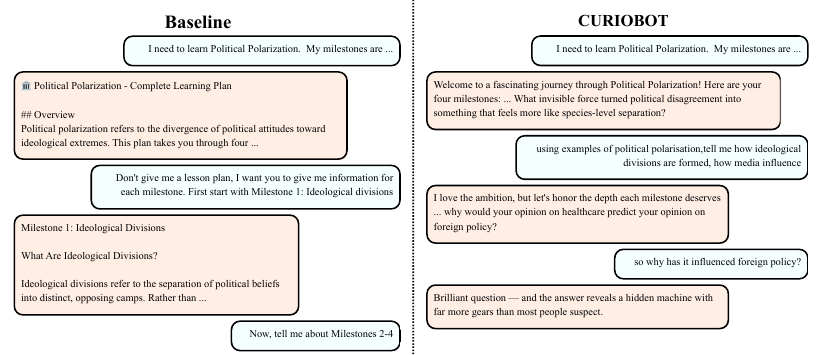}
    \caption{Example conversation on the topic Political Polarization.}
    \label{fig:example-conv-2}
\end{figure}

Figure~\ref{fig:example-conv-1} and Figure~\ref{fig:example-conv-2} show more examples how \system sustains a conversation.

\section{Study-oriented Mode Scores} \label{appendix:study-mode-scores}

Table~\ref{tab:study-mode-scores} shows study mode scores comparison to Baseline and \systemnospace. The $p$-value columns show the statistical significance of mean difference between \system and study mode. 

\begin{table*}[!h]
  \centering
    \resizebox{\textwidth}{!}{
    \begin{tabular}{ccccccccc}
      \hline
      \textbf{Dimension} & \multicolumn{4}{c}{\textbf{Gemini}} & \multicolumn{4}{c}{\textbf{Claude}} \\
      \cmidrule(lr){2-5} \cmidrule(lr){6-9} 
      & \textbf{Baseline} & \textbf{\system} & \textbf{Study} & \textbf{$p$-value} & \textbf{Baseline} & \textbf{\system} & \textbf{Study} & \textbf{$p$-value}  \\
      & & & & \systemnospace-Study & & & & \systemnospace-Study \\
      \hline
      $L_1$ & $ 2.51 \pm 0.15 $ & $ 3.21 \pm 0.12 $ & $ 2.90 \pm 0.10 $ & \textbf{**} & $ 2.72 \pm 0.11 $ & $ 3.29 \pm 0.12 $ & $ 2.53 \pm 0.10 $ & \textbf{**} \\
      $L_2$ & $ 1.94 \pm 0.11 $ & $ 2.86 \pm 0.10 $ & $ 3.47 \pm 0.08 $ & ** & $ 2.06 \pm 0.10 $ & $ 3.02 \pm 0.11 $ & $ 3.72 \pm 0.10 $ & ** \\
      $L_3$ & $ 2.56 \pm 0.14 $ & $ 3.36 \pm 0.12 $ & $ 3.43 \pm 0.09 $ & n.s. & $ 2.88 \pm 0.11 $ & $ 3.63 \pm 0.12 $ & $ 3.43 \pm 0.10 $ & n.s. \\
      $L_4$ & $ 2.55 \pm 0.14 $ & $ 3.40 \pm 0.12 $ & $ 3.61 \pm 0.09 $ & n.s. & $ 2.73 \pm 0.11 $ & $ 3.54 \pm 0.10 $ & $ 3.47 \pm 0.10 $ & n.s. \\
      \hline
      $T_1$ & $ 5.30 \pm 0.10 $ & $ 3.66 \pm 0.10 $ & $ 5.33 \pm 0.05 $ & ** & $ 5.28 \pm 0.11 $ & $ 5.33 \pm 0.06 $ & $ 5.64 \pm 0.06 $ & ** \\
      $T_2$ & $ 2.79 \pm 0.11 $ & $ 2.80 \pm 0.10 $ & $ 4.69 \pm 0.08 $ & ** & $ 3.07 \pm 0.13 $ & $ 4.19 \pm 0.10 $ & $ 5.14 \pm 0.09 $ & ** \\
      $T_3$ & $ 4.42 \pm 0.13 $ & $ 3.25 \pm 0.11 $ & $ 5.28 \pm 0.06 $ & ** & $ 3.88 \pm 0.16 $ & $ 4.58 \pm 0.07 $ & $ 5.47 \pm 0.06 $ & ** \\
      $T_4$ & $ 3.88 \pm 0.14 $ & $ 3.59 \pm 0.11 $ & $ 5.02 \pm 0.07 $ & ** & $ 3.82 \pm 0.12 $ & $ 4.76 \pm 0.09 $ & $ 5.35 \pm 0.08 $ & ** \\
      $T_5$ & $ 4.07 \pm 0.11 $ & $ 4.25 \pm 0.09 $ & $ 4.36 \pm 0.06 $ & n.s. & $ 3.85 \pm 0.13 $ & $ 5.27 \pm 0.06 $ & $ 4.48 \pm 0.07 $ & \textbf{**} \\
      $T_6$ & $ 3.13 \pm 0.12 $ & $ 3.18 \pm 0.10 $ & $ 5.36 \pm 0.07 $ & ** & $ 2.94 \pm 0.12 $ & $ 4.35 \pm 0.10 $ & $ 5.63 \pm 0.09 $ & ** \\
      \hline
    \end{tabular}
    }
  \caption{\label{tab:study-mode-scores}Dimension-wise scores aggregated over domains and complexities. Notation for $p$-value of statistical significance of mean difference: $*$ and $**$ represent $p < 0.05 $ and $ p < 0.005$ respectively.}
\end{table*}

\section{LLM-as-a-Judge Scores for Each Judge Model} \label{appendix:3-judges}

This section reports the LLM-as-a-Judge results disaggregated by judge model, with scores averaged over three independent runs. Across all three judges a consistent self-preference effect emerges on instructional quality ($T_1$): each judge ranks its own Baseline above its own \systemnospace, although the gap is statistically insignificant in every case. We unpack the per-judge findings below.

\subsection{Gemini-as-a-Judge}

Table~\ref{tab:judge-gemini} reports the dimension-wise scores from \texttt{gemini-3-flash-preview}, aggregated over domains and complexities. Gains for \system are uniform: every learner-side dimension ($L_1$--$L_4$) and every tutor-side dimension other than $T_1$ improves over the Baseline for all three LLMs, with the majority of differences significant at $p < 0.005$. The sole exception is the self-preference on $T_1$, where the Gemini judge rates Gemini's Baseline ($4.49$) marginally above Gemini's \system ($4.19$, n.s.); for Claude and GPT as LLMs, the same judge still records strong $T_1$ gains for \system ($p < 0.005$).

\begin{table*}[!h]
  \centering
  \resizebox{\textwidth}{!}{
    \begin{tabular}{cccccccccc}
      \hline
      \textbf{Dimension} & \multicolumn{3}{c}{\textbf{Gemini}} & \multicolumn{3}{c}{\textbf{Claude}} & \multicolumn{3}{c}{\textbf{GPT}} \\
      \cmidrule(lr){2-4} \cmidrule(lr){5-7} \cmidrule(lr){8-10}
      & \textbf{Baseline} & \textbf{\system} & \textbf{$p$-value} & \textbf{Baseline} & \textbf{\system} & \textbf{$p$-value} & \textbf{Baseline} & \textbf{\system} & \textbf{$p$-value} \\
      \hline
      $L_1$ & $ 2.35 \pm 0.23 $ & $ 2.86 \pm 0.18 $ & \textbf{*} & $ 2.60 \pm 0.19 $ & $ 3.03 \pm 0.17 $ & n.s. & $ 2.20 \pm 0.18 $ & $ 3.12 \pm 0.19 $ & \textbf{**} \\
      $L_2$ & $ 1.91 \pm 0.21 $ & $ 2.81 \pm 0.19 $ & \textbf{**} & $ 1.91 \pm 0.16 $ & $ 3.23 \pm 0.20 $ & \textbf{**} & $ 1.91 \pm 0.16 $ & $ 2.66 \pm 0.21 $ & \textbf{**} \\
      $L_3$ & $ 2.26 \pm 0.23 $ & $ 3.09 \pm 0.16 $ & \textbf{**} & $ 2.62 \pm 0.20 $ & $ 3.42 \pm 0.17 $ & \textbf{**} & $ 2.35 \pm 0.18 $ & $ 3.10 \pm 0.18 $ & \textbf{**} \\
      $L_4$ & $ 2.51 \pm 0.24 $ & $ 3.37 \pm 0.17 $ & \textbf{**} & $ 2.70 \pm 0.18 $ & $ 3.62 \pm 0.17 $ & \textbf{**} & $ 2.38 \pm 0.19 $ & $ 3.23 \pm 0.17 $ & \textbf{**} \\
      \hline
      $T_1$ & $ 4.49 \pm 0.18 $ & $ 4.19 \pm 0.18 $ & n.s. & $ 4.10 \pm 0.17 $ & $ 5.12 \pm 0.10 $ & \textbf{**} & $ 3.95 \pm 0.13 $ & $ 4.75 \pm 0.11 $ & \textbf{**} \\
      $T_2$ & $ 2.31 \pm 0.21 $ & $ 3.01 \pm 0.21 $ & \textbf{**} & $ 2.20 \pm 0.18 $ & $ 3.91 \pm 0.19 $ & \textbf{**} & $ 1.68 \pm 0.17 $ & $ 3.07 \pm 0.20 $ & \textbf{**} \\
      $T_3$ & $ 3.32 \pm 0.29 $ & $ 4.21 \pm 0.18 $ & \textbf{*} & $ 2.32 \pm 0.18 $ & $ 4.63 \pm 0.17 $ & \textbf{**} & $ 2.80 \pm 0.20 $ & $ 4.75 \pm 0.13 $ & \textbf{**} \\
      $T_4$ & $ 3.48 \pm 0.29 $ & $ 4.09 \pm 0.24 $ & \textbf{*} & $ 3.38 \pm 0.20 $ & $ 4.75 \pm 0.17 $ & \textbf{**} & $ 3.00 \pm 0.21 $ & $ 4.37 \pm 0.17 $ & \textbf{**} \\
      $T_5$ & $ 3.40 \pm 0.23 $ & $ 4.44 \pm 0.18 $ & \textbf{**} & $ 2.73 \pm 0.22 $ & $ 5.01 \pm 0.13 $ & \textbf{**} & $ 1.97 \pm 0.16 $ & $ 4.07 \pm 0.13 $ & \textbf{**} \\
      $T_6$ & $ 3.00 \pm 0.24 $ & $ 3.88 \pm 0.19 $ & \textbf{**} & $ 2.43 \pm 0.21 $ & $ 4.57 \pm 0.21 $ & \textbf{**} & $ 2.48 \pm 0.20 $ & $ 4.18 \pm 0.17 $ & \textbf{**} \\
      \hline
    \end{tabular}
  }
  \caption{\label{tab:judge-gemini}Dimension-wise scores from \texttt{gemini-3-flash-preview} aggregated over domains and complexities. Notation for $p$-value of statistical significance of mean difference: $*$ and $**$ represent $p < 0.05 $ and $ p < 0.005$ respectively.}
\end{table*}

\subsection{Claude-as-a-Judge}

Table~\ref{tab:judge-claude} reports the scores assigned by \texttt{claude-sonnet-4-6}. Learner-side improvements ($L_1$--$L_4$) remain consistent and highly significant ($p < 0.005$) across all three LLMs, mirroring the pattern observed under the Gemini judge. The tutor-side picture is more nuanced. For Claude and GPT as LLMs, \system improves over the Baseline on nearly every tutor-side dimension, typically at $p < 0.005$; the self-preference on $T_1$ surfaces here as a null result, with the judge rating its own Baseline and \system nearly identically ($5.79$ vs.\ $5.69$, n.s.). For Gemini as the LLM, however, the judge systematically favours the Baseline: $T_1$, $T_3$, and $T_5$ show significant Baseline advantages, while $T_2$, $T_4$, and $T_6$ are statistically indistinguishable. This asymmetry indicates that, among the three judges, Claude is the least favourable to \system when scoring a different-family model on the tutor dimensions.

\begin{table*}[!h]
  \centering
  \resizebox{\textwidth}{!}{
    \begin{tabular}{cccccccccc}
      \hline
      \textbf{Dimension} & \multicolumn{3}{c}{\textbf{Gemini}} & \multicolumn{3}{c}{\textbf{Claude}} & \multicolumn{3}{c}{\textbf{GPT}} \\
      \cmidrule(lr){2-4} \cmidrule(lr){5-7} \cmidrule(lr){8-10}
      & \textbf{Baseline} & \textbf{\system} & \textbf{$p$-value} & \textbf{Baseline} & \textbf{\system} & \textbf{$p$-value} & \textbf{Baseline} & \textbf{\system} & \textbf{$p$-value} \\
      \hline
      $L_1$ & $ 2.02 \pm 0.16 $ & $ 2.68 \pm 0.10 $ & \textbf{**} & $ 2.19 \pm 0.10 $ & $ 2.81 \pm 0.11 $ & \textbf{**} & $ 2.27 \pm 0.16 $ & $ 2.88 \pm 0.17 $ & \textbf{**} \\
      $L_2$ & $ 1.70 \pm 0.12 $ & $ 2.42 \pm 0.11 $ & \textbf{**} & $ 1.73 \pm 0.13 $ & $ 2.65 \pm 0.17 $ & \textbf{**} & $ 1.72 \pm 0.12 $ & $ 2.44 \pm 0.13 $ & \textbf{**} \\
      $L_3$ & $ 2.10 \pm 0.17 $ & $ 2.75 \pm 0.14 $ & \textbf{**} & $ 2.36 \pm 0.11 $ & $ 3.06 \pm 0.15 $ & \textbf{**} & $ 2.31 \pm 0.14 $ & $ 2.88 \pm 0.14 $ & \textbf{**} \\
      $L_4$ & $ 1.85 \pm 0.14 $ & $ 2.58 \pm 0.12 $ & \textbf{**} & $ 1.96 \pm 0.09 $ & $ 2.84 \pm 0.13 $ & \textbf{**} & $ 2.06 \pm 0.13 $ & $ 2.68 \pm 0.15 $ & \textbf{**} \\
      \hline
      $T_1$ & $ 5.79 \pm 0.09 $ & $ 3.40 \pm 0.12 $ & ** & $ 5.79 \pm 0.06 $ & $ 5.69 \pm 0.07 $ & n.s. & $ 5.17 \pm 0.07 $ & $ 5.50 \pm 0.05 $ & \textbf{**} \\
      $T_2$ & $ 2.81 \pm 0.16 $ & $ 2.68 \pm 0.12 $ & n.s. & $ 3.01 \pm 0.15 $ & $ 4.53 \pm 0.13 $ & \textbf{**} & $ 2.49 \pm 0.13 $ & $ 3.86 \pm 0.13 $ & \textbf{**} \\
      $T_3$ & $ 4.73 \pm 0.09 $ & $ 2.75 \pm 0.11 $ & ** & $ 4.04 \pm 0.12 $ & $ 4.67 \pm 0.08 $ & \textbf{**} & $ 4.44 \pm 0.10 $ & $ 4.77 \pm 0.07 $ & \textbf{**} \\
      $T_4$ & $ 3.32 \pm 0.18 $ & $ 3.11 \pm 0.13 $ & n.s. & $ 3.25 \pm 0.13 $ & $ 4.65 \pm 0.14 $ & \textbf{**} & $ 3.00 \pm 0.15 $ & $ 4.37 \pm 0.15 $ & \textbf{**} \\
      $T_5$ & $ 4.35 \pm 0.14 $ & $ 3.88 \pm 0.15 $ & * & $ 4.14 \pm 0.10 $ & $ 5.48 \pm 0.06 $ & \textbf{**} & $ 3.06 \pm 0.09 $ & $ 4.90 \pm 0.08 $ & \textbf{**} \\
      $T_6$ & $ 2.91 \pm 0.15 $ & $ 2.88 \pm 0.12 $ & n.s. & $ 2.62 \pm 0.13 $ & $ 4.43 \pm 0.15 $ & \textbf{**} & $ 2.63 \pm 0.11 $ & $ 3.85 \pm 0.12 $ & \textbf{**} \\
      \hline
    \end{tabular}
  }
  \caption{\label{tab:judge-claude}Dimension-wise scores from \texttt{claude-sonnet-4-6} aggregated over domains and complexities. Notation for $p$-value of statistical significance of mean difference: $*$ and $**$ represent $p < 0.05 $ and $ p < 0.005$ respectively.}
\end{table*}

\subsection{GPT-as-a-Judge}

Table~\ref{tab:judge-gpt} reports the scores produced by \texttt{gpt-5}. As with the other two judges, learner-side dimensions improve significantly with \system across all LLMs. On the tutor-side, the self-preference on $T_1$ surfaces as a non-significant Baseline lead for the judge's own model ($5.60$ vs.\ $5.41$), whereas the same dimension shows a significant Baseline advantage when the LLM is Gemini ($5.62$ vs.\ $3.41$, $p < 0.005$) or Claude ($5.96$ vs.\ $5.17$, $p < 0.005$). The Baseline bias extends beyond $T_1$: for Gemini as the LLM, the GPT judge rates the Baseline higher on $T_2$, $T_3$, $T_4$, and $T_6$ (all $p < 0.005$), with only $T_5$ statistically indistinguishable; for Claude as the LLM, $T_3$ similarly favours the Baseline. On GPT as the LLM, by contrast, \system wins on $T_2$, $T_4$, $T_5$, and $T_6$. The overall pattern is that this judge is the most generous to Baselines on cross-family comparisons while still rewarding \system on its own model's learner dimensions.

\begin{table*}[!h]
  \centering
  \resizebox{\textwidth}{!}{
    \begin{tabular}{cccccccccc}
      \hline
      \textbf{Dimension} & \multicolumn{3}{c}{\textbf{Gemini}} & \multicolumn{3}{c}{\textbf{Claude}} & \multicolumn{3}{c}{\textbf{GPT}} \\
      \cmidrule(lr){2-4} \cmidrule(lr){5-7} \cmidrule(lr){8-10}
      & \textbf{Baseline} & \textbf{\system} & \textbf{$p$-value} & \textbf{Baseline} & \textbf{\system} & \textbf{$p$-value} & \textbf{Baseline} & \textbf{\system} & \textbf{$p$-value} \\
      \hline
      $L_1$ & $ 3.16 \pm 0.33 $ & $ 4.09 \pm 0.19 $ & \textbf{**} & $ 3.40 \pm 0.17 $ & $ 4.01 \pm 0.24 $ & n.s. & $ 2.78 \pm 0.21 $ & $ 3.95 \pm 0.23 $ & \textbf{**} \\
      $L_2$ & $ 2.22 \pm 0.21 $ & $ 3.53 \pm 0.19 $ & \textbf{**} & $ 2.53 \pm 0.20 $ & $ 3.17 \pm 0.18 $ & \textbf{*} & $ 1.96 \pm 0.15 $ & $ 2.92 \pm 0.22 $ & \textbf{**} \\
      $L_3$ & $ 3.33 \pm 0.25 $ & $ 4.25 \pm 0.18 $ & \textbf{**} & $ 3.65 \pm 0.17 $ & $ 4.42 \pm 0.18 $ & \textbf{*} & $ 2.75 \pm 0.17 $ & $ 4.16 \pm 0.20 $ & \textbf{**} \\
      $L_4$ & $ 3.30 \pm 0.22 $ & $ 4.26 \pm 0.15 $ & \textbf{**} & $ 3.53 \pm 0.14 $ & $ 4.16 \pm 0.15 $ & \textbf{*} & $ 2.97 \pm 0.14 $ & $ 4.00 \pm 0.18 $ & \textbf{**} \\
      \hline
      $T_1$ & $ 5.62 \pm 0.07 $ & $ 3.41 \pm 0.15 $ & ** & $ 5.96 \pm 0.05 $ & $ 5.17 \pm 0.11 $ & ** & $ 5.60 \pm 0.06 $ & $ 5.41 \pm 0.11 $ & n.s. \\
      $T_2$ & $ 3.26 \pm 0.17 $ & $ 2.70 \pm 0.12 $ & ** & $ 3.99 \pm 0.20 $ & $ 4.17 \pm 0.16 $ & n.s. & $ 2.94 \pm 0.15 $ & $ 3.47 \pm 0.19 $ & \textbf{*} \\
      $T_3$ & $ 5.20 \pm 0.04 $ & $ 2.78 \pm 0.11 $ & ** & $ 5.27 \pm 0.17 $ & $ 4.43 \pm 0.12 $ & ** & $ 5.44 \pm 0.08 $ & $ 5.11 \pm 0.12 $ & * \\
      $T_4$ & $ 4.83 \pm 0.11 $ & $ 3.57 \pm 0.12 $ & ** & $ 4.84 \pm 0.14 $ & $ 4.88 \pm 0.13 $ & n.s. & $ 4.68 \pm 0.08 $ & $ 5.04 \pm 0.13 $ & \textbf{**} \\
      $T_5$ & $ 4.46 \pm 0.10 $ & $ 4.42 \pm 0.12 $ & n.s. & $ 4.69 \pm 0.09 $ & $ 5.31 \pm 0.08 $ & \textbf{**} & $ 3.72 \pm 0.08 $ & $ 4.64 \pm 0.11 $ & \textbf{**} \\
      $T_6$ & $ 3.47 \pm 0.17 $ & $ 2.78 \pm 0.10 $ & ** & $ 3.77 \pm 0.19 $ & $ 4.05 \pm 0.14 $ & n.s. & $ 3.23 \pm 0.10 $ & $ 3.67 \pm 0.16 $ & \textbf{**} \\
      \hline
    \end{tabular}
  }
  \caption{\label{tab:judge-gpt}Dimension-wise scores from \texttt{gpt-5} aggregated over domains and complexities. Notation for $p$-value of statistical significance of mean difference: $*$ and $**$ represent $p < 0.05 $ and $ p < 0.005$ respectively.}
\end{table*}

\section{Scores Across Domains and Complexities}\label{appendix:scores-complexity-domain}


\begin{figure*}[!h]
    \centering
    \begin{subfigure}[b]{0.3\textwidth}
        \includegraphics[width=\textwidth]{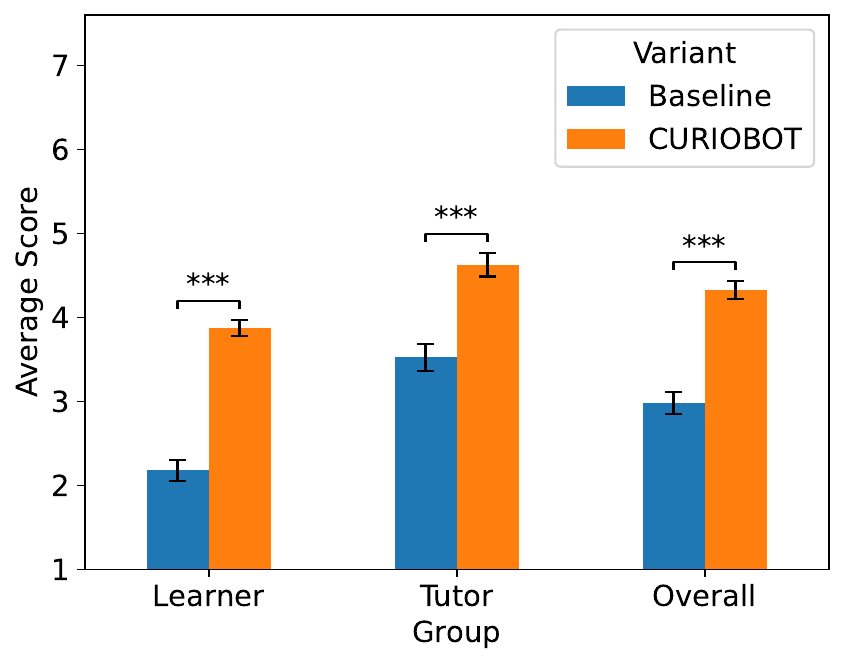}
        \caption{Low SS}\label{fig:low-ss}
    \end{subfigure}
    \hfill
    \begin{subfigure}[b]{0.3\textwidth}
        \includegraphics[width=\textwidth]{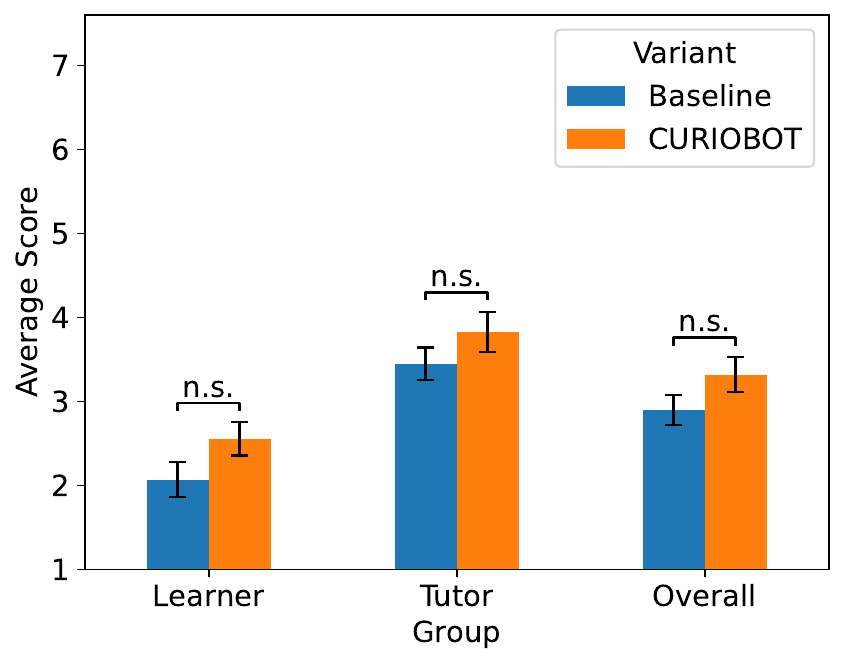}
        \caption{Low AH}\label{fig:low-ah}
    \end{subfigure}
    \hfill
    \begin{subfigure}[b]{0.3\textwidth}
        \includegraphics[width=\textwidth]{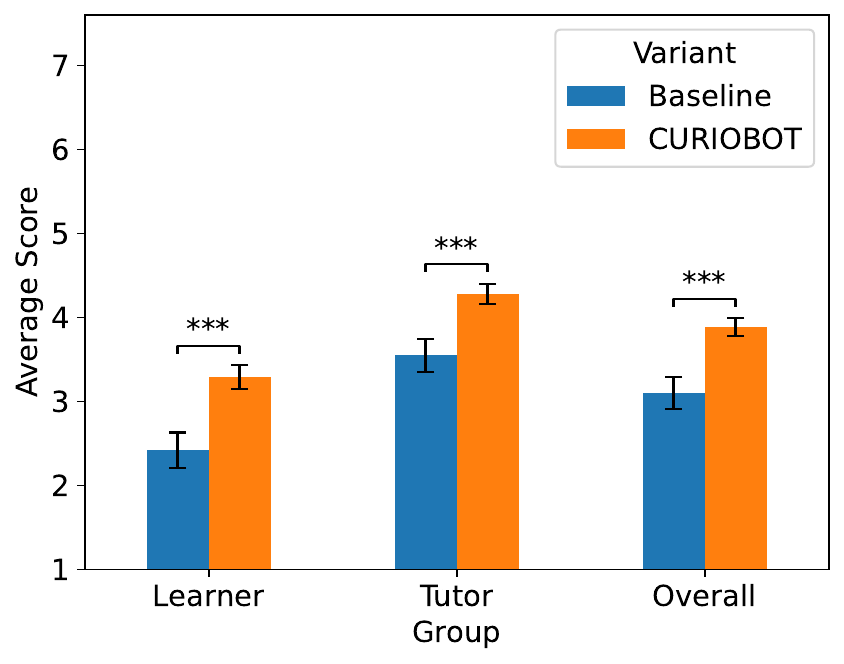}
        \caption{Low STEM}\label{fig:low-stem}
    \end{subfigure}
    
    \vspace{0.5em}
    \begin{subfigure}[b]{0.3\textwidth}
        \includegraphics[width=\textwidth]{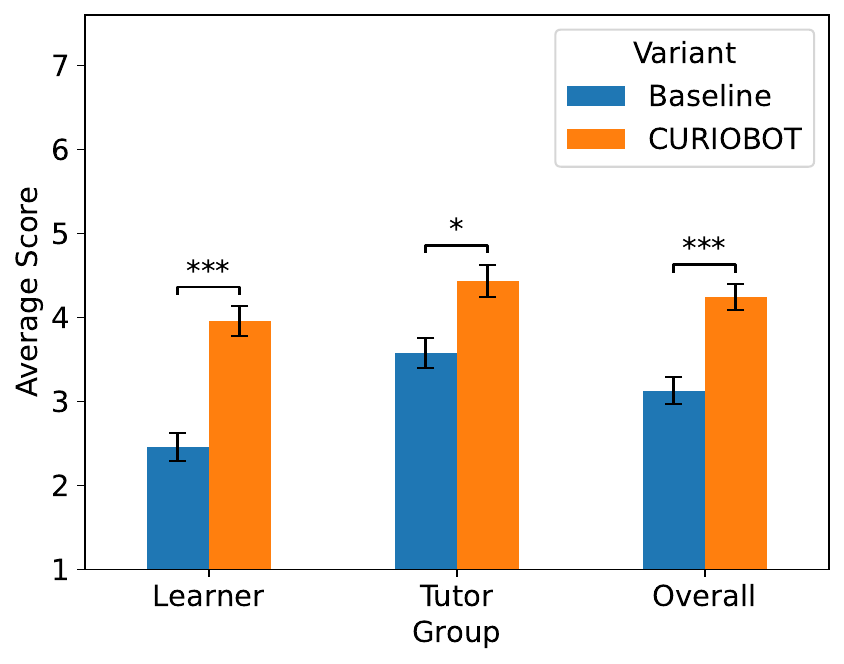}
        \caption{Medium SS}\label{fig:medium-ss}
    \end{subfigure}
    \hfill
    \begin{subfigure}[b]{0.3\textwidth}
        \includegraphics[width=\textwidth]{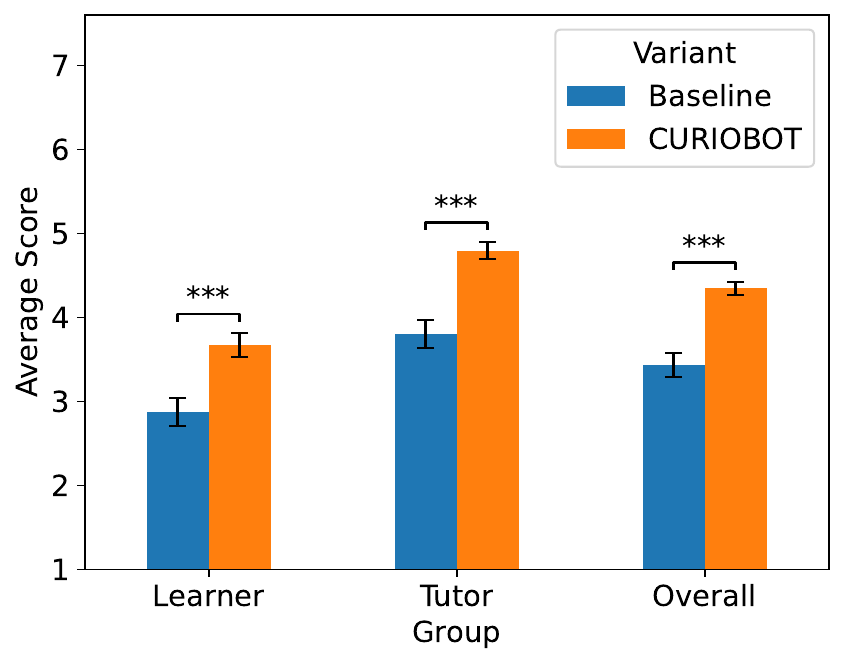}
        \caption{Medium AH}\label{fig:medium-ah}
    \end{subfigure}
    \hfill
    \begin{subfigure}[b]{0.3\textwidth}
        \includegraphics[width=\textwidth]{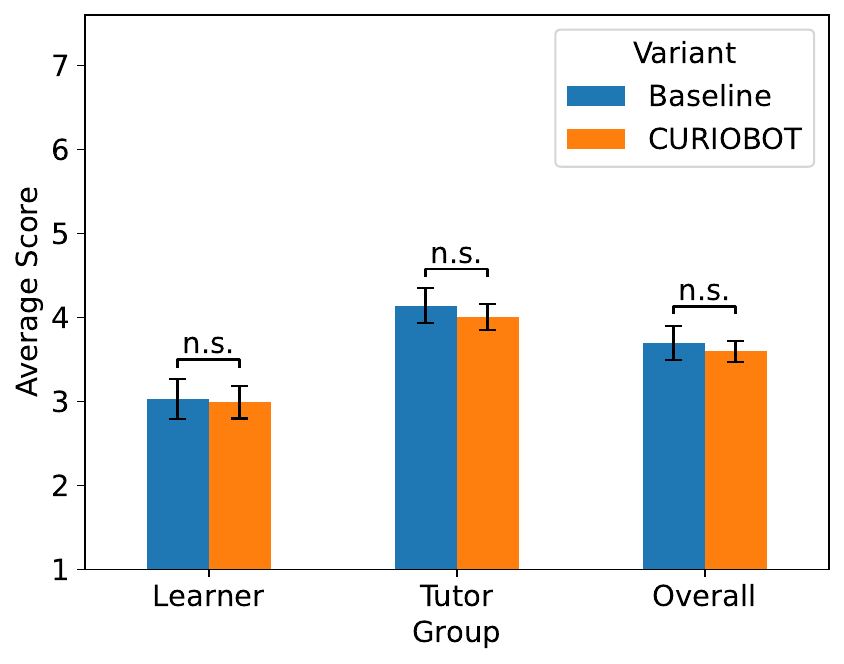}
        \caption{Medium STEM}\label{fig:medium-stem}
    \end{subfigure}
    
    \vspace{0.5em}
    \begin{subfigure}[b]{0.3\textwidth}
        \includegraphics[width=\textwidth]{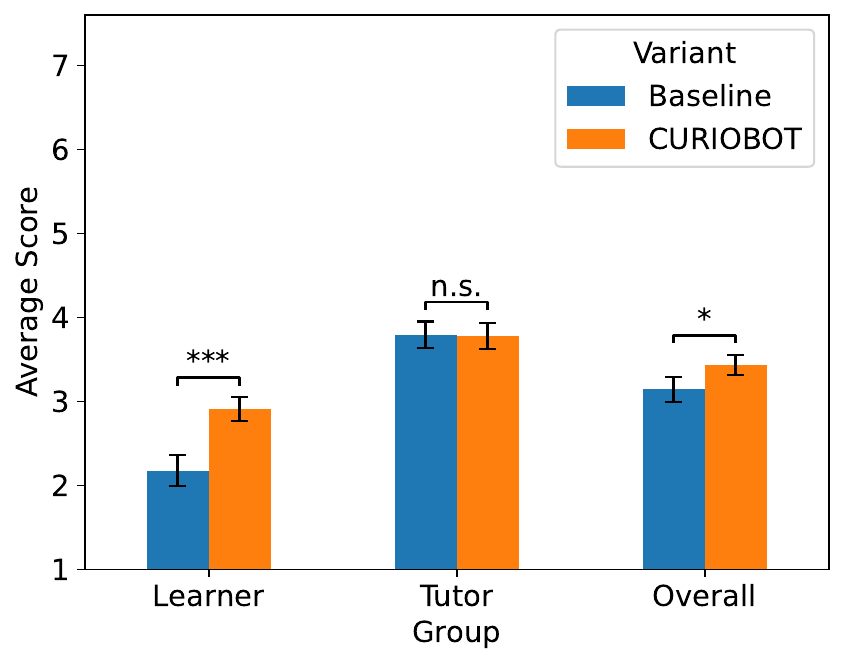}
        \caption{High SS}\label{fig:high-ss}
    \end{subfigure}
    \hfill
    \begin{subfigure}[b]{0.3\textwidth}
        \includegraphics[width=\textwidth]{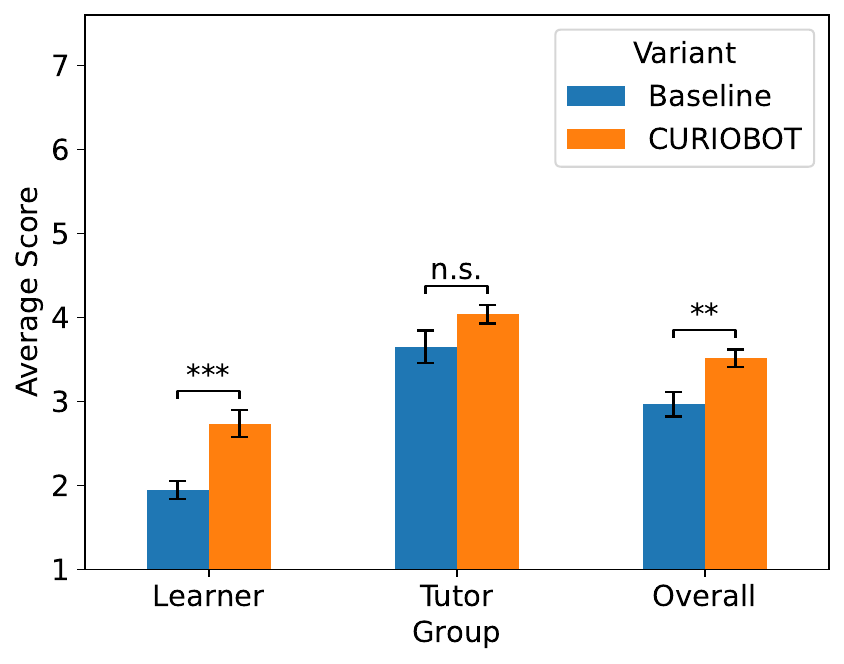}
        \caption{High AH}\label{fig:high-ah}
    \end{subfigure}
    \hfill
    \begin{subfigure}[b]{0.3\textwidth}
        \includegraphics[width=\textwidth]{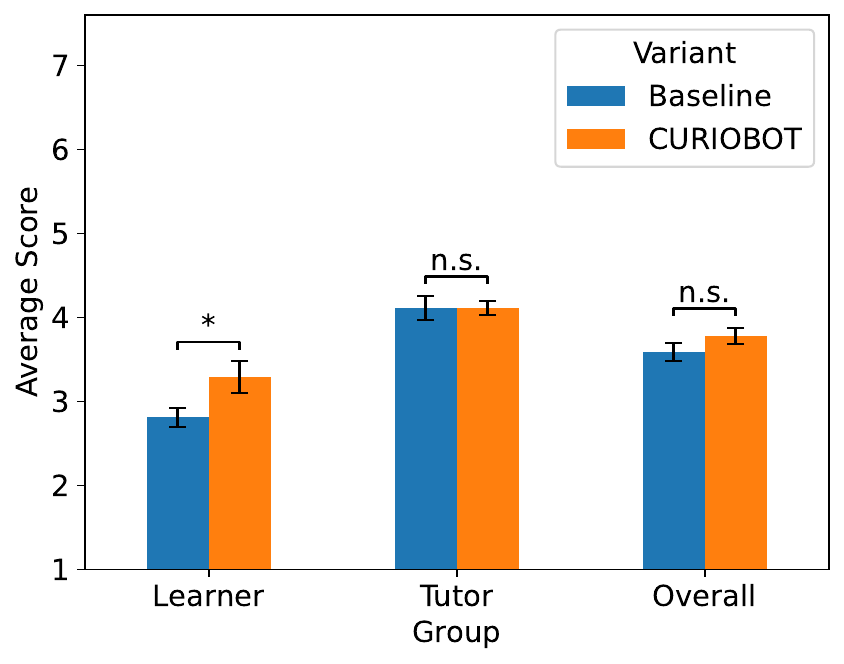}
        \caption{High STEM}\label{fig:high-stem}
    \end{subfigure}
    
    \caption{Aggregated results over LLMs; columns represents subject areas while rows represents complexity levels. Notation for $p$-value of statistical significance of mean difference: $*$ and $**$ represent $p < 0.05 $ and $ p < 0.005$ respectively.}
    \label{fig:complexity-vs-domain}
\end{figure*}

\twocolumn

%% file: ref.bib
@article{gruber2014states,
  title={States of curiosity modulate hippocampus-dependent learning via the dopaminergic circuit},
  author={Gruber, Matthias J and Gelman, Bernard D and Ranganath, Charan},
  journal={Neuron},
  volume={84},
  number={2},
  pages={486--496},
  year={2014},
  publisher={Elsevier}
}

@article{ten2021humans,
  title={Humans monitor learning progress in curiosity-driven exploration},
  author={Ten, Alexandr and Kaushik, Pramod and Oudeyer, Pierre-Yves and Gottlieb, Jacqueline},
  journal={Nature communications},
  volume={12},
  number={1},
  pages={5972},
  year={2021},
  publisher={Nature Publishing Group UK London}
}

@article{berlyne1960conflict,
  title={Conflict, arousal, and curiosity.},
  author={Berlyne, Daniel E},
  year={1960},
  publisher={McGraw-Hill Book Company}
}

@article{berlyne1962motivational,
  title={Motivational problems raised by exploratory and epistemic behavior.},
  author={Berlyne, Daniel E},
  year={1962},
  publisher={McGraw-Hill}
}

@inproceedings{papineni2002bleu,
  title={Bleu: a method for automatic evaluation of machine translation},
  author={Papineni, Kishore and Roukos, Salim and Ward, Todd and Zhu, Wei-Jing},
  booktitle={Proceedings of the 40th annual meeting of the Association for Computational Linguistics},
  pages={311--318},
  year={2002}
}

@article{zhang2019bertscore,
  title={Bertscore: Evaluating text generation with bert},
  author={Zhang, Tianyi and Kishore, Varsha and Wu, Felix and Weinberger, Kilian Q and Artzi, Yoav},
  journal={arXiv preprint arXiv:1904.09675},
  year={2019}
}

@inproceedings{maurya2025unifying,
  title={Unifying AI tutor evaluation: An evaluation taxonomy for pedagogical ability assessment of LLM-powered AI tutors},
  author={Maurya, Kaushal Kumar and Srivatsa, Kv Aditya and Petukhova, Kseniia and Kochmar, Ekaterina},
  booktitle={Proceedings of the 2025 Conference of the Nations of the Americas Chapter of the Association for Computational Linguistics: Human Language Technologies (Volume 1: Long Papers)},
  pages={1234--1251},
  year={2025}
}

@inproceedings{pauzi2025automating,
  title={Automating pedagogical evaluation of LLM-based conversational agents},
  author={Pauzi, Zaki and Dodman, Michael and Mavrikis, Manolis},
  booktitle={Ceur Workshop Proceedings},
  volume={4006},
  year={2025},
  organization={CEUR}
}

@article{tack2022ai,
  title={The AI teacher test: Measuring the pedagogical ability of blender and GPT-3 in educational dialogues},
  author={Tack, Ana{\"\i}s and Piech, Chris},
  journal={arXiv preprint arXiv:2205.07540},
  year={2022}
}

@inproceedings{macina2023mathdial,
  title={Mathdial: A dialogue tutoring dataset with rich pedagogical properties grounded in math reasoning problems},
  author={Macina, Jakub and Daheim, Nico and Chowdhury, Sankalan and Sinha, Tanmay and Kapur, Manu and Gurevych, Iryna and Sachan, Mrinmaya},
  booktitle={Findings of the Association for Computational Linguistics: EMNLP 2023},
  pages={5602--5621},
  year={2023}
}

@inproceedings{daheim2024stepwise,
  title={Stepwise verification and remediation of student reasoning errors with large language model tutors},
  author={Daheim, Nico and Macina, Jakub and Kapur, Manu and Gurevych, Iryna and Sachan, Mrinmaya},
  booktitle={Proceedings of the 2024 Conference on Empirical Methods in Natural Language Processing},
  pages={8386--8411},
  year={2024}
}

@inproceedings{wang2024bridging,
  title={Bridging the novice-expert gap via models of decision-making: A case study on remediating math mistakes},
  author={Wang, Rose and Zhang, Qingyang and Robinson, Carly and Loeb, Susanna and Demszky, Dorottya},
  booktitle={Proceedings of the 2024 Conference of the North American Chapter of the Association for Computational Linguistics: Human Language Technologies (Volume 1: Long Papers)},
  pages={2174--2199},
  year={2024}
}

@article{qian2025dean,
  title={Dean of llm tutors: exploring comprehensive and automated evaluation of llm-generated educational feedback via llm feedback evaluators},
  author={Qian, Keyang and Cheng, Yixin and Guan, Rui and Dai, Wei and Jin, Flora and Yang, Kaixun and Nawaz, Sadia and Swiecki, Zachari and Chen, Guanliang and Yan, Lixiang and others},
  journal={arXiv preprint arXiv:2508.05952},
  year={2025}
}

@article{zheng2023judging,
  title={Judging llm-as-a-judge with mt-bench and chatbot arena},
  author={Zheng, Lianmin and Chiang, Wei-Lin and Sheng, Ying and Zhuang, Siyuan and Wu, Zhanghao and Zhuang, Yonghao and Lin, Zi and Li, Zhuohan and Li, Dacheng and Xing, Eric and others},
  journal={Advances in neural information processing systems},
  volume={36},
  pages={46595--46623},
  year={2023}
}

@inproceedings{pal2024autotutor,
  title={Autotutor meets large language models: A language model tutor with rich pedagogy and guardrails},
  author={Pal Chowdhury, Sankalan and Zouhar, Vil{\'e}m and Sachan, Mrinmaya},
  booktitle={Proceedings of the Eleventh ACM Conference on Learning@ Scale},
  pages={5--15},
  year={2024}
}

@article{mollick2023assigning,
  title={Assigning AI: Seven approaches for students, with prompts},
  author={Mollick, Ethan and Mollick, Lilach},
  journal={arXiv preprint arXiv:2306.10052},
  year={2023}
}

@article{denny2024computing,
  title={Computing education in the era of generative AI},
  author={Denny, Paul and Prather, James and Becker, Brett A and Finnie-Ansley, James and Hellas, Arto and Leinonen, Juho and Luxton-Reilly, Andrew and Reeves, Brent N and Santos, Eddie Antonio and Sarsa, Sami},
  journal={Communications of the ACM},
  volume={67},
  number={2},
  pages={56--67},
  year={2024},
  publisher={ACM New York, NY, USA}
}

@inproceedings{favero2024enhancing,
  title={Enhancing critical thinking in education by means of a Socratic chatbot},
  author={Favero, Lucile and P{\'e}rez-Ortiz, Juan Antonio and K{\"a}ser, Tanja and Oliver, Nuria},
  booktitle={International workshop on AI in education and educational research},
  pages={17--32},
  year={2024},
  organization={Springer}
}

@article{bonino2024euler,
  title={EULER: Fine-Tuning a Large Language Model for Socratic Interactions.},
  author={Bonino, Giulia and Sanmartino, Gabriele and Pinheiro, Giovanni Gatti and Papotti, Paolo and Troncy, Rapha{\"e}l and Michiardi, Pietro},
  journal={AIxEDU@ AI* IA},
  volume={3879},
  year={2024}
}

@article{team2024learnlm,
  title={Learnlm: Improving gemini for learning},
  author={LearnLM Team and Modi, Abhinit and Veerubhotla, Aditya Srikanth and others},
  journal={arXiv preprint arXiv:2412.16429},
  year={2024}
}

@article{liu2024socraticlm,
  title={SocraticLM: Exploring socratic personalized teaching with large language models},
  author={Liu, Jiayu and Huang, Zhenya and Xiao, Tong and Sha, Jing and Wu, Jinze and Liu, Qi and Wang, Shijin and Chen, Enhong},
  journal={Advances in Neural Information Processing Systems},
  volume={37},
  pages={85693--85721},
  year={2024}
}

@article{kidd2015psychology,
  title={The psychology and neuroscience of curiosity},
  author={Kidd, Celeste and Hayden, Benjamin Y},
  journal={Neuron},
  volume={88},
  number={3},
  pages={449--460},
  year={2015},
  publisher={Elsevier}
}

@article{frank2025cognitive,
  title={Cognitive modeling using artificial intelligence},
  author={Frank, Michael C and Goodman, Noah D},
  journal={Annual Review of Psychology},
  volume={77},
  year={2025},
  publisher={Annual Reviews}
}

@article{binz2023using,
  title={Using cognitive psychology to understand GPT-3},
  author={Binz, Marcel and Schulz, Eric},
  journal={Proceedings of the National Academy of Sciences},
  volume={120},
  number={6},
  pages={e2218523120},
  year={2023},
  publisher={National Academy of Sciences}
}

@misc{hagendorff2023human,
  title={Human-like intuitive behavior and reasoning biases emerged in large language models but disappeared in ChatGPT. Nature Computational Science, 3 (10), 833-838},
  author={Hagendorff, Thilo and Fabi, Sarah and Kosinski, Michal},
  year={2023}
}

@article{argyle2023out,
  title={Out of one, many: Using language models to simulate human samples},
  author={Argyle, Lisa P and Busby, Ethan C and Fulda, Nancy and Gubler, Joshua R and Rytting, Christopher and Wingate, David},
  journal={Political Analysis},
  volume={31},
  number={3},
  pages={337--351},
  year={2023},
  publisher={Cambridge University Press}
}

@article{trott2024can,
  title={Can large language models help augment English psycholinguistic datasets?},
  author={Trott, Sean},
  journal={Behavior Research Methods},
  volume={56},
  number={6},
  pages={6082--6100},
  year={2024},
  publisher={Springer}
}

@inproceedings{qian2025generating,
  title={Generating Neurolinguistic Stimuli Using LLM Prompting},
  author={Qian, Ming and Patten, Terry and Lynn, Spencer and Winder, Aaron and Pickering, Maxwell},
  booktitle={International Conference on Human-Computer Interaction},
  pages={404--419},
  year={2025},
  organization={Springer}
}

@inproceedings{lu2026mind,
  title={Mind the gap: The divergence between human and llm-generated tasks},
  author={Lu, Yi-Long and Song, Jiajun and Zhang, Chunhui and Wang, Wei},
  booktitle={Proceedings of the AAAI Conference on Artificial Intelligence},
  volume={40},
  number={3},
  pages={1964--1972},
  year={2026}
}

@inproceedings{laverghetta2023generating,
  title={Generating better items for cognitive assessments using large language models},
  author={Laverghetta Jr, Antonio and Licato, John},
  booktitle={Proceedings of the 18th workshop on innovative use of NLP for building educational applications (BEA 2023)},
  pages={414--428},
  year={2023}
}

@article{lee2022curious,
  title={Do curious students learn more science in an immersive virtual reality environment? Exploring the impact of advance organizers and epistemic curiosity},
  author={Lee, Silvia Wen-Yu and Hsu, Ying-Tai and Cheng, Kun-Hung},
  journal={Computers \& Education},
  volume={182},
  pages={104456},
  year={2022},
  publisher={Elsevier}
}

@article{eren2009examining,
  title={Examining the Relationship between Epistemic Curiosity and Achievement Goals.},
  author={Eren, Altay},
  journal={Eurasian Journal of Educational Research (EJER)},
  number={36},
  year={2009}
}

@article{sarac2022does,
  title={Does E-learning Trigger Epistemic Curiosity?},
  author={SARAC, Seda and Enisa, MEDE and AKGUN, Ergun},
  journal={Journal of Qualitative Research in Education},
  number={30},
  year={2022}
}

@article{decaro2015achievement,
  title={Achievement motivation and knowledge development during exploratory learning},
  author={DeCaro, Daniel A and DeCaro, Marci S and Rittle-Johnson, Bethany},
  journal={Learning and Individual Differences},
  volume={37},
  pages={13--26},
  year={2015},
  publisher={Elsevier}
}

@article{kapur2008productive,
  title={Productive failure},
  author={Kapur, Manu},
  journal={Cognition and instruction},
  volume={26},
  number={3},
  pages={379--424},
  year={2008},
  publisher={Taylor \& Francis}
}

@inproceedings{deng2024towards,
  title={Towards human-centered proactive conversational agents},
  author={Deng, Yang and Liao, Lizi and Zheng, Zhonghua and Yang, Grace Hui and Chua, Tat-Seng},
  booktitle={Proceedings of the 47th International ACM SIGIR Conference on Research and Development in Information Retrieval},
  pages={807--818},
  year={2024}
}
